%% file: CoRL_main.tex
\documentclass{article}
\usepackage{amssymb}
\usepackage{times}
\usepackage{multicol}
\usepackage{multirow}
\usepackage[bookmarks=true]{hyperref}
\usepackage{amsmath}
\usepackage{amsthm}
\usepackage{graphicx}
\usepackage[dvipsnames,table,xcdraw]{xcolor}
\usepackage[inkscapelatex=false]{svg}
\usepackage[font=small,labelfont=bf]{caption}
\usepackage{subcaption}
\usepackage{bbding}
\usepackage{soul}
\usepackage{cancel}
\usepackage[normalem]{ulem}
\usepackage{booktabs}
\usepackage{soul}
\usepackage{algorithmic}
\usepackage{algorithm}
\usepackage{wrapfig}

\usepackage{booktabs}
\usepackage{colortbl}
\usepackage[table]{xcolor}

\usepackage{siunitx}
\newcommand{\numerr}[2]{#1 $\pm$ #2}

\newcommand*{\draft}{} 
\input{src/comments}
\input{src/commands}

\usepackage[final]{corl_2025} 

\title{Generative Visual Foresight Meets Task-Agnostic Pose Estimation in Robotic Table-Top Manipulation}

\author{
\textbf{Chuye Zhang}$^{*1}$ \And \textbf{Xiaoxiong Zhang}\thanks{Equal contribution, order by dice rolling.} $^{\ 1}$\And \textbf{Wei Pan}$^{1}$ \And \textbf{Linfang Zheng}$^{\dagger 2, 3}$ \And \textbf{Wei Zhang}\thanks{The corresponding authors.}$^{\ 1, 2}$ \and  \vspace{-5px} \\ 
$^{1}$Southern University of Science and Technology \\ $^{2}$LimX Dynamics \and $^{3}$The University of Hong Kong \\
{\tt\small$\{$\text{12110807}, \text{12433017}, \text{12211810}$\}$@mail.sustech.edu.cn}\\
{\tt\small lfzheng@hku.hk, zhangw3@sustech.edu.cn} \\
}

\begin{document}
\maketitle

\vspace{-28px}
\begin{figure}[h]
    \centering
    \includegraphics[width=0.90\columnwidth,trim={0.4cm 0.9cm 0.3cm 0.4cm},clip]{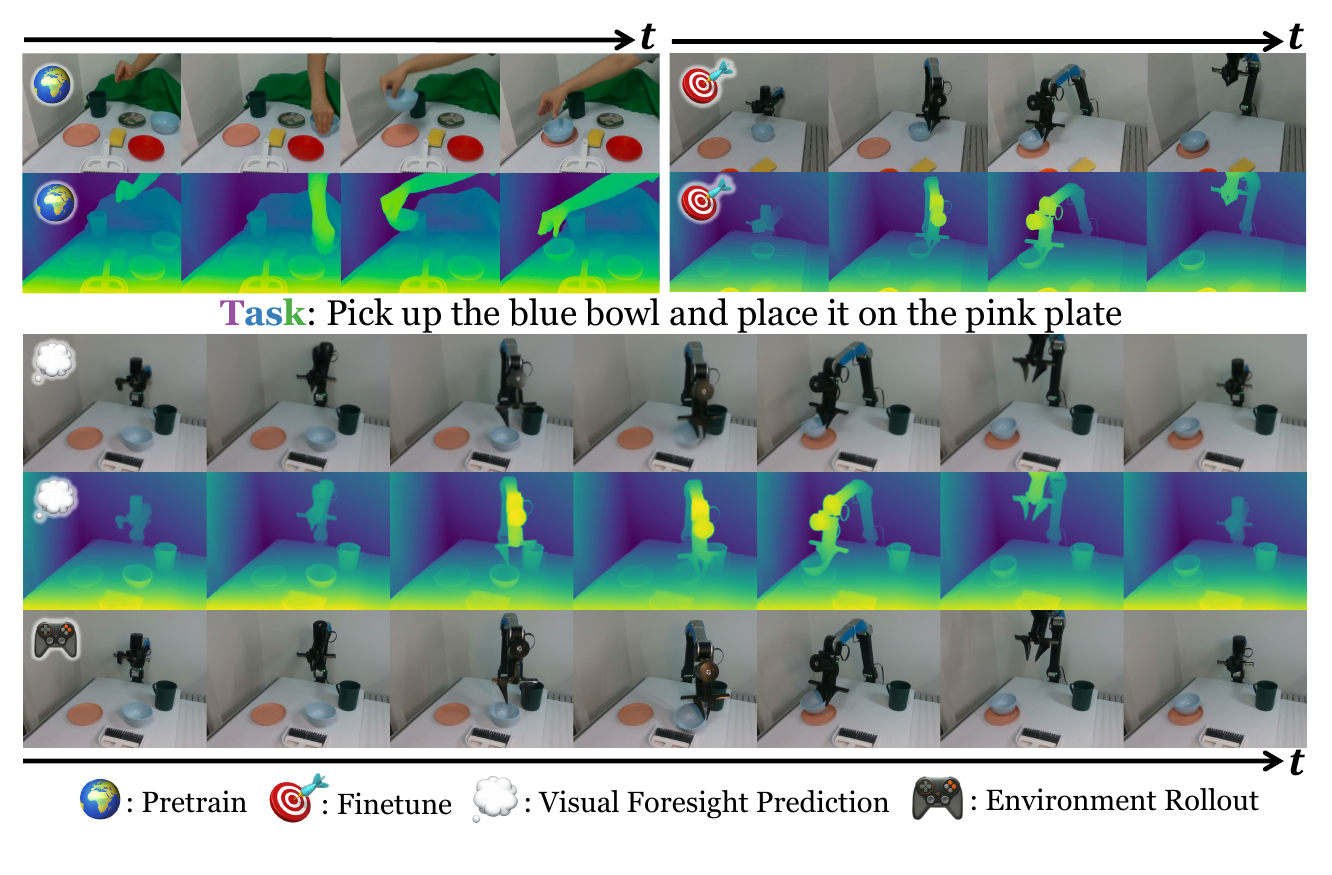}
    \vspace{-3px}
    \caption{\footnotesize{\textbf{High-level illustration of GVF-TAPE.}
    Given a single RGB observation and a task description, GVF-TAPE predicts future RGB-D frames via a generative foresight model. A decoupled pose estimator then extracts end-effector poses, enabling closed-loop manipulation without action labels.
    }
    }
    \vspace{-4mm}
\end{figure}


\input{tex/abstract}



\input{tex/intro}

\input{tex/related}

\input{tex/method}
\input{tex/experiments}
\input{tex/conclusion}

\input{tex/future_works}




	

\acknowledgments{This work was supported by the Guangdong Science and Technology Program under Grant No.~2024B1212010002.
}


\newpage
\bibliography{references}  

\newpage
\input{tex/appendix}

\end{document}

%% file: src/comments.tex
\ifdefined\draft
    \usepackage{comment}
    \usepackage[textsize=footnotesize]{todonotes}

    \DeclareRobustCommand{\chuye}[1]
    {{\todo[color=blue!40,inline]{CZ: #1}}}

    \usepackage[nopar=false]{lipsum}
    
    \DeclareRobustCommand{\fake}[1]{\color{gray!40!}{\lipsum[][#1]}\color{black}}
    
\else 
\usepackage{pbalance} 

\newcommand{\chuye}[1]{}
\newcommand{\fake}[1]{}

\fi

%% file: src/commands.tex
\newcommand{\stkout}[1]{\ifmmode\text{\sout{\ensuremath{#1}}}\else\sout{#1}\fi}


%% file: tex/abstract.tex
\begin{abstract}

{ Robotic manipulation in unstructured environments requires systems that can generalize across diverse tasks while maintaining robust and reliable performance. We introduce {GVF-TAPE}, a closed-loop framework that combines generative visual foresight with task-agnostic pose estimation to enable scalable robotic manipulation. GVF-TAPE employs a generative video model to predict future RGB-D frames from a single side-view RGB image and a task description, offering visual plans that guide robot actions. A decoupled pose estimation model then extracts end-effector poses from the predicted frames, translating them into executable commands via low-level controllers. By iteratively integrating video foresight and pose estimation in a closed loop, GVF-TAPE achieves real-time, adaptive manipulation across a broad range of tasks. Extensive experiments in both simulation and real-world settings demonstrate that our approach reduces reliance on task-specific action data and generalizes effectively, providing a practical and scalable solution for intelligent robotic systems. The
video and code can be found at \url{https://clearlab-sustech.github.io/gvf-tape/}}.
\end{abstract}
\vspace{-3mm}

%% file: tex/intro.tex
\section{Introduction}

 {Humans develop an intuitive understanding of hand kinematics through continuous interaction with their environment~\citep{Pilacinski2024natural, Dieter2014seehands}. Studies have highlighted the strong sensory coupling between vision and body awareness~\citep{Faivre2015_body_cons, Yokosaka2015_vis_compress}, enabling people to predict the visual consequences of their actions before executing them. Inspired by this capability, we propose to enable robots to \textbf{imagine future visual scenes and infer their end-effector states to guide actions}. This insight motivates the design of \textbf{GVF-TAPE} (\textbf{G}enerative \textbf{V}isual \textbf{F}oresight with \textbf{T}ask-\textbf{A}gnostic \textbf{P}ose \textbf{E}stimation), a closed-loop framework that combines generative video prediction with task-agnostic pose estimation to achieve scalable, real-time robotic manipulation.}

Recent advances in robotic manipulation leverage large-scale, vision-language-action models~\citep{kim2024openvla, octo_2023, brohan2023rt1roboticstransformerrealworld, brohan2023rt2visionlanguageactionmodelstransfer,black2024pi0visionlanguageactionflowmodel}. However, scaling such models is challenging due to the cost and effort required for human-annotated demonstrations. To address this, action-free datasets have gained increasing attention. {Some approaches learn general representations for policy learning~\citep{r3m, jiang2024robotspretrain} or label action-free dataset with latent action~\citep{ye2024lapa}, }while others guide actions by predicting intermediate visual cues such as future frames~\citep{black2023susie, du2023unipi,clover}, point tracks~\citep{wen2024atm,xu2024im2flw2act} and sphere pose~\citep{genima}.
{{Despite these advances, many methods still depend on task-specific action supervision during downstream learning or require rigid setups, limiting their scalability and adaptability. Recent efforts to eliminate action labels through dense correspondence~\citep{Ko2023avdc}, goal-conditioned exploration~\citep{luo2024groundingv2a}, or stereo-based pose estimation~\citep{liang2024dreamitaterealworldvisuomotorpolicy} have shown promise, but often face challenges related to real-world flexibility, data collection efficiency, or closed-loop deployment.} These limitations motivate the need for a \textbf{task-agnostic}, \textbf{action-label-free framework} that can plan through future visual prediction and execute actions reliably in real-time, without relying on specialized hardware or task-specific supervision.

{In this work, we introduce \textbf{GVF-TAPE}, a novel video-based framework for robotic manipulation that decouples the phases of visual planning and action execution. Our approach leverages a \textbf{generative video model} to predict future RGB-D frames from a single side-view RGB image and a task description, providing a rich visual plan for decision making. A \textbf{task-agnostic pose estimation model} then extracts 6-DoF end-effector poses from the generated frames and translates them into executable actions through low-level controllers via inverse kinematics. Crucially, the pose estimation model is trained solely on random exploration data, making it simple to collect and scalable across different robots and environments. By integrating video foresight and task-agnostic pose estimation in a closed-loop system, GVF-TAPE enables robust, real-time manipulation across a wide range of tasks. Extensive experiments in both simulation and real-world settings demonstrate that our method matches or outperforms prior video-pretrained, action-labeled, and self-exploration-based approaches while requiring significantly less task-specific data.}

{The main contributions of this work are:
\begin{itemize}
\item We propose GVF-TAPE, a closed-loop, action-label-free framework that combines generative visual foresight and task-agnostic pose estimation for real-time robotic manipulation.
\item We develop a scalable training pipeline by leveraging random exploration data for pose learning and large-scale video pretraining for foresight, eliminating the need for expert-labeled demonstrations.
\item We demonstrate that GVF-TAPE achieves real-time deployment in both simulation and real-world environments, and significantly outperforms prior action-labeled, video-pretrained, and self-exploration-based methods across diverse manipulation tasks.
\end{itemize}}

%% file: tex/related.tex
\section{Related Work}
\textbf{Visual foresight for robotic manipulation.} The research on visual foresight models have become a hotspot for robotic manipulation by using it as auxiliary loss, guidance feature or sub-goal. \citep{du2023unipi, Ko2023avdc, black2023susie, wen2024atm, xu2024im2flw2act, liang2024dreamitaterealworldvisuomotorpolicy, shridhar2024generativeimageactionmodels, tian2024seer, gr2}. Approaches like \citep{tian2024seer,gr1, gr2} integrate visual foresight as auxiliary loss for policy learning to obtain better dynamics comprehension.  Methods like \citep{wen2024atm,xu2024im2flw2act, shridhar2024generativeimageactionmodels, du2023unipi, black2023susie} choose to train a model that generates temporal feature like point track \citep{wen2024atm, xu2024im2flw2act}, sphere pose \cite{shridhar2024generativeimageactionmodels} or sub-goal image \citep{du2023unipi, black2023susie, clover} to guide the policy learning. While these methods exploited visual foresight to enhance policy learning, they still rely on action-labeled data to train a inverse dynamic model mapping visual foresight to executable action. Methods like \citep{Ko2023avdc, luo2024groundingv2a, liang2024dreamitaterealworldvisuomotorpolicy} bridged this gap by eliminating the need for action-labeled data. AVDC~\citep{Ko2023avdc} uses optical flow and dense matching to obtain action, suffering from manipulation precision and dependence on the robot mask. V2A~\citep{luo2024groundingv2a} obtains an inverse dynamic model by self-exploration and bootsrapping, facing challenges in task specificity, data-efficiency and real-world safety constraints. Dreamitate~\citep{liang2024dreamitaterealworldvisuomotorpolicy} estimates the pose of the robotic arm in the predicted video, which requires manipulator CAD model, stereo setup and camera calibration, as well as struggles in inference time and precision. Our research aims to develop an agile and close-loop video prediction and execution framework, additionally it's easy-to-obtain, less-dependent and practical.

\textbf{Pose Estimation in Robotics} 
Pose estimation has been extensively studied in both Computer Vision and Robotics. Object pose estimation can generally be categorized into instance-level pose estimation, which requires CAD models~\citep{SLAM6D_2022_CVPR, BB8_Rad_2017_ICCV, G2L_Net, PVN3D_CVPR_2019, Tremblay_2018_Bbox_Corners, ZebraPose_2022_CVPR, DPOD_2019_ICCV, OneShot_2d3d_2022_CVPR, Epro-PnP_2022_CVPR, SurfEmb_2022_CVPR, SSS_6D_2018_CVPR, DFTr_Zhou_2023_ICCV,zheng2022TPAE}, and category-level pose estimation, which generalizes across object instances within a given category~\citep{SPD_eccv_2020, UDA-COPE_2022_CVPR, CASS_2020_CVPR, ShaPO_ECCV_2022, zheng2024georefgeometricalignmentshape, hs-pose}. In the context of tabletop manipulators and articulated robots, important pose estimation tasks include object poses, end-effector poses, and joint angles (1D poses) ~\citep{Ko2023avdc, huang2024rekep, zuo2019craves}
Keypoint-based approaches have also been widely adopted for estimating 6D camera-to-robot poses and joint angles~\citep{DREAM, lu2024ctrnetxcameratorobotposeestimation, dspdh_2024, zuo2019craves, tian2024robokeygen}. Additionally, ~\citep{labbé2021singleviewrobojoint} introduced a render-and-compare method that overlays articulated CAD models for pose estimation. Foundation pose model ~\citep{wen2024foundationposeunified6dpose} create a general and easy-to-adopt workflow for object pose estimation. However, the constrain to articulated objects makes it less convenient in estimating gripper aperture. FEEPE ~\citep{wu2025foundationfeaturedrivenonlineendeffector} construct a training free foundation model for robot end effector pose estimation. However, we focus on end-to-end robot-centric approach which utilize extensive proprioception data which are more adaptable for certain manipulator embodiment. Similarly, ~\citep{deepeef} also use deep neural network for end effector pose estimation. Keeping the advantages of end-to-end approaches, it needs additional ground truth depth information.  In contrast to these approaches, our work employs a lightweight, end-to-end deep learning model that requires only easily and automatically collected random exploration data, and synthesized depth generated by ready-to-use Video Depth Anything Model~\citep{video_depth_anything},  making it both efficient and practical for real-world robotic applications.

%% file: tex/method.tex
\section{Method}
\subsection{Problem Formulation}
\label{sec:problem_formulation}
Our goal is to develop a closed-loop robotic manipulation system for tabletop environments that combines visual foresight with task-agnostic pose estimation. Given a single side-view RGB observation $x_0$ of the scene and a task description $c$, the system predicts a sequence of future robot actions in the form of end-effector poses. Specifically, the system generates a pose trajectory $\mathcal{T} = {T_1, T_2, \dots, T_h}$, where each $T_i = (\mathbf{p}_i, \mathbf{q}_i, g_i)$ consists of the 3D position $\mathbf{p}_i \in \mathbb{R}^3$, the orientation $\mathbf{q}_i \in \mathbb{R}^4$ represented as a unit quaternion, and the gripper state $g_i \in [0,1]$ indicating the gripper opening. Thus, we aim to learn a mapping function $f: (x_0, c) \rightarrow \mathcal{T}$ that allows the robot to execute tasks robustly while continuously adapting to dynamic environments.

\subsection{Framework Overview}
We propose GVF-TAPE (Generative Visual Foresight and Task-Agnostic Pose Estimation model), a decoupled two-stage framework for closed-loop robotic manipulation, as illustrated in Fig.~\ref{fig: system_overview}. GVF-TAPE plans directly in the visual space by first predicting future observations and then inferring the corresponding end-effector poses through task-agnostic pose estimation. This design enables greater generalization and eliminates the need for expert demonstrations.

\subsection{ Text-Conditioned Visual Foresight for RGB-D Prediction}
\label{sec:video-generation}

{Our visual foresight module predicts future RGB-D frames conditioned on the current RGB observation $x_0$ and a task description $c$. Prior models~\citep{du2023unipi, Ko2023avdc} predict only RGB frames, while CLOVER~\citep{clover} generates RGB-D but requires explicit depth inputs, limiting scalability. In contrast, our approach infers depth implicitly, enabling training and deployment without depth sensors. Using off-the-shelf depth estimators~\citep{video_depth_anything}, we also enable pretraining on large-scale RGB-only datasets.} 

{To model the conditional distribution $p(x_{1:h} \mid x_0, c)$ efficiently, we adopt rectified flow~\citep{rectified_flow, opensora}, which transforms an initial noisy sequence $x^1_{1:h} \sim \mathcal{N}(0, I)$ toward a clean video prediction $x^0_{1:h}$ following:}
\begin{equation}
    dx^t_{1:h}=(x^0_{1:h}-x^1_{1:h})dt,
    \label{rectified-flow}
\end{equation}
where $x^t_{1:h}$ interpolates between noise and ground truth. The velocity model $v_\theta$ is trained to predict the displacement between the noisy and clean sequences by minimizing: 
\begin{equation}
    \mathcal{L}=\|v_\theta(x^t_{1:h},x_0,c,t)-(x^0_{1:h}-x^1_{1:h})\|^2, t\sim U(0,1).
    \label{rectified-flow-loss}
\end{equation}
Here, $x^t_{1:h}$ is a linear interpolation between $x^0_{1:h}$ and $x^1_{1:h}$, given by $x^t_{1:h} = t x^1_{1:h} + (1 - t) x^0_{1:h}$, with $t$ denoting the noise level.
\input{images/tex/framework_overview}

Since future prediction requires modeling both spatial and temporal dynamics, we adopt a lightweight 3D U-Net~\citep{Ko2023avdc} as the backbone for the velocity model $v_\theta$, and encode the task description $c$ using a CLIP text encoder~\citep{clip}. This architecture enables efficient and scalable visual foresight for real-time robotic planning.

\subsection{Task-Agnostic Pose Estimation Model}
\label{sec:ik_impl}
To translate the generated video frames into executable robot actions, we employ a task-agnostic pose estimation model. Unlike previous methods~\citep{du2023unipi, black2023susie, Ko2023avdc} that rely on inverse dynamics and temporal dependencies, our approach processes each frame independently, improving flexibility and generalization across different tasks.

Given an RGB image $x_i$ and its corresponding depth map from the foresight model, the pose estimator $\pi_{\phi}$ predicts the end-effector pose $T_i = (\mathbf{p}_i, \mathbf{q}_i, g_i)$, as defined in Section~\ref{sec:problem_formulation}. The model is trained to minimize a Smooth L1 loss:
\begin{equation}
\mathcal{L}=\text{SmoothL1}(\pi_{\phi}(x_i)-T_i).
\end{equation}
To fuse RGB and depth information effectively, we adopt two separated pretrained ViT-base encoders~\citep{DBLP:journals/corr/abs-2010-11929} for RGB and depth modalities, and apply a multi-head cross-attention mechanism: 
\begin{equation}
\mathbf{f}_{\text{fused}} = \mathcal{A} \, (\mathbf{Q} = \mathbf{d}_{\text{cls}},\ \mathbf{K} = \mathbf{r}_{\text{tok}},\ \mathbf{V} = \mathbf{r}_{\text{tok}}),
\end{equation}
{where $\mathbf{f}_{\text{fused}}$ denotes the fused feature representation, and $\mathcal{A}$ is the multi-head attention module. The query $\mathbf{Q}$ is the CLS token from the depth encoder ($\mathbf{d}_{\text{cls}}$), while the keys and values ($\mathbf{K}, \mathbf{V}$) are patch tokens from the RGB image ($\mathbf{r}_{\text{tok}}$), including its CLS token $\mathbf{r}{\text{cls}}$. Further architectural details are provided in the Supplementary Material.}

{To collect training data for the pose estimation model, we use a random exploration strategy: sampling $T_i = (\mathbf{p}_i, \mathbf{q}_i, g_i)$ uniformly within a predefined workspace range. An off-the-shelf controller drives the robot to each sampled pose, with safety constraints enforced in real-world settings. Details are included in the Supplementary Material.}  

%% file: images/tex/framework_overview.tex
\begin{figure*}[!tbp]
    \centering
    \includegraphics[width=0.95\columnwidth,trim={2.2cm 7.5cm 0.8cm 5.62cm},clip]{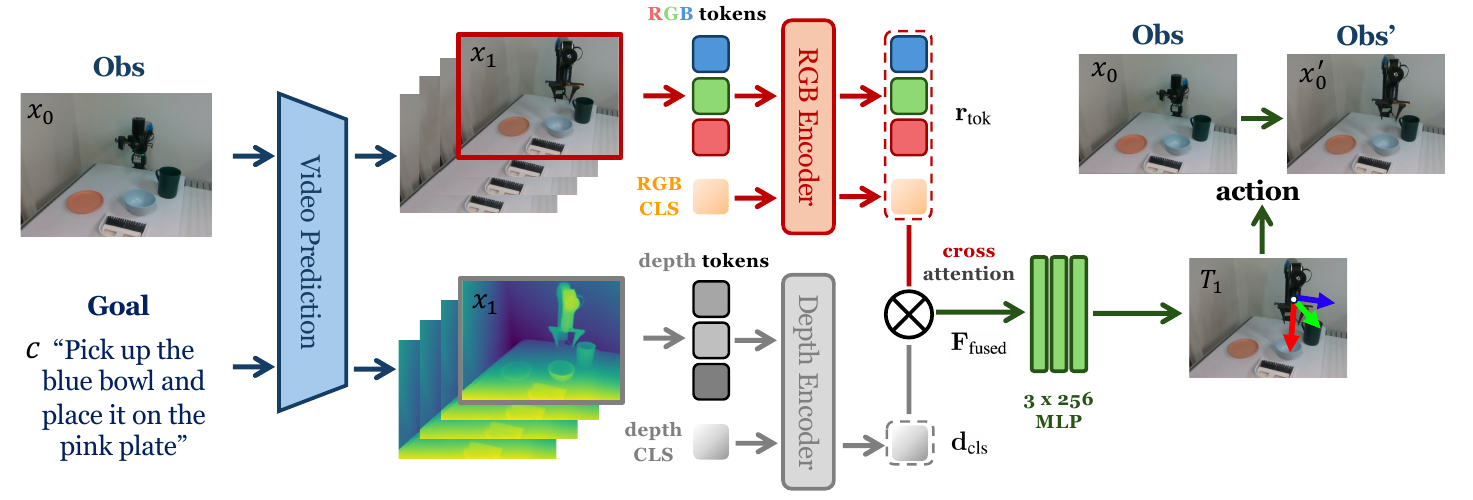}
    \vspace{-4px}
    \caption{\footnotesize{
         \textbf{Framework Overview.}  
GVF-TAPE first generates a future RGB-D video conditioned on the current RGB observation and task description. A transformer-based pose estimation model then extracts the end-effector pose from each predicted frame and sends it to a low-level controller for execution. After completing the predicted trajectory, the system receives a new observation and repeats the process in a closed-loop manner.}}
        
    \label{fig: system_overview}
    \vspace{-4mm}
\end{figure*}

%% file: tex/experiments.tex
\vspace{-3px}
\section{Experiment}
\label{Experiment}
\vspace{-2px}
We evaluate GVF-TAPE through extensive simulation and real-world experiments to answer the following key questions: (1) How does our approach, trained on random exploration data, compare with state-of-the-art video pre-training imitation learning methods that require action labels? (2) How does it compare with other video prediction methods that map future or goal images to actions? (3) Can GVF-TAPE benefit from pre-train on external video data (\textit{e.g.}, human manipulation videos)? (4) How effective are our design choices, such as rectified flow and depth inference?

\paragraph{Datasets.} We evaluate GVF-TAPE in both simulated and real-world settings. In simulation, we adopt the LIBERO benchmark~\citep{liu2023libero}—a suite of language-conditioned manipulation tasks designed for benchmarking generalizable robotic agents. Detailed information about LIBERO is provided in Sec.\ref{sec:overview of libero}. For training the task-agnostic pose estimation model, we generate over 400k RGB-D/pose pairs per task suite via simulation. The datasets used to train the video generation model in simulation are described in Sec.\ref{Compare with video predicton method} and Sec.\ref{Compare with video pre-training method}.
For real-world experiments, we collect 18k RGB-D/pose pairs through random exploration (Sec.\ref{sec:ik_impl}) for pose estimation, and acquire 20 teleoperated demonstrations per task to train the video generation model.
\paragraph{Baselines.} We compare GVF-TAPE against two categories of prior work. First, we evaluate against video pretraining methods including \textit{R3M-finetune}~\citep{r3m}, \textit{VPT}~\citep{baker2022videopretrainingvptlearning}, \textit{UniPi}~\citep{du2023unipi}, and \textit{ATM}~\citep{wen2024atm}, all of which rely on action-labeled demonstrations for policy learning. Second, we compare with video prediction-based approaches such as \textit{DP}~\citep{chi2023diffusionpolicy}, \textit{GCDP}~\citep{luo2024groundingv2a}, \textit{AVDC}~\citep{Ko2023avdc}, \textit{SuSIE}~\citep{black2023susie}, and \textit{V2A}~\citep{luo2024groundingv2a}, which learn from videos by predicting intermediate visual representations or sub-goals. Notably, unlike V2A, our method does not require goal-conditioned exploration and can be trained entirely offline. Baseline training and evaluation protocols follow those reported in~\citep{wen2024atm,luo2024groundingv2a}.

\input{tables/libero_main}

\subsection{Comparison with Video Pre-training Methods}
\label{Compare with video pre-training method}
\vspace{-4px}
To evaluate the effectiveness of our proposed method, we compare our method with state-of-the-art video pre-training imitation learning approaches~\citep{r3m,du2023unipi,wen2024atm,baker2022videopretrainingvptlearning} on LIBERO-spatial, LIBERO-object, and LIBERO-goal, covering a total of 30 language-conditioned manipulation tasks. For baselines, each task is trained with 50 video demonstrations and 10 action-labeled trajectories. 

The results, presented in Table~\ref{tab:libero-main}, show that our method (GVF-TAPE) outperforms all baselines requiring action-labeled data in LIBERO-spatial and LIBERO-object, achieving 27.00\% and 18.70\% performance gains, respectively. In LIBERO-goal, our method ranks second, being 11.03\% lower than the ATM. Upon further analysis, we found that tasks in LIBERO-goal often require precise manipulation in gripper occluded scenes, such as opening drawers.\input{images/tex/difficult-scenes}Our current setup uses a single fixed camera, which can limit visibility of fine-grained interactions (see Fig.~\ref{fig:difficult-scenes} (c–d)). Incorporating wrist-mounted or multi-view inputs may help mitigate this limitation, which we leave for future work.

Despite these challenges, GVF-TAPE achieves the best average performance across all suites, surpassing ATM by 11.56\%. These results show that our framework can achieve competitive or superior performance compared to action-labeled methods, without requiring any expert actions, and highlight its potential for scalable and label-free robot learning.

\paragraph{Data Efficiency}
Reducing reliance on large quantities of robot data is increasingly important due to the cost of human teleoperation and annotation. \input{images/tex/pretrain_sr} To evaluate GVF-TAPE’s data efficiency, we pretrain the video generation model on LIBERO-90 and fine-tune it on LIBERO-Spatial, LIBERO-Object, and LIBERO-Goal using 20\%, 50\%, and 90\% of available task data. We assess both video generation quality, using LPIPS and SSIM metrics, and downstream task performance, comparing against models trained from scratch. As shown in Fig.~\ref{fig:pretrain_scratch_and_ddim_flow}, the pre-trained model consistently outperforms the scratch model across all data proportions, demonstrating that pretraining on external video sources improves video fidelity. For task success, as depicted in  Fig.~\ref{fig:pre_train_sr} GVF-TAPE achieves 68\% with only 20\% of demonstration data (10 demonstrations per task), and further improves to 77\% when pretrained on LIBERO-90, surpassing the previous state-of-the-art by 5.43\%. These results highlight the strong data efficiency and transferability of our approach.

\input{tables/libero-v2a}

\vspace{-2px}
\subsection{Comparison with Video Prediction Methods}
\vspace{-4px}
\label{Compare with video predicton method}

 We further compare GVF-TAPE against video prediction-based approaches~\citep{luo2024groundingv2a, Ko2023avdc, black2023susie} on eight tasks from two living room scenes in the LIBERO-100 suite, following the evaluation protocol in~\citep{luo2024groundingv2a}. Each method is trained on 20 video demonstrations per task (160 total). Baselines include DP~\citep{chi2023diffusionpolicy}, GCDP~\citep{luo2024groundingv2a}, and SuSIE~\citep{black2023susie}, which rely on action-labeled data, and V2A~\citep{luo2024groundingv2a} and AVDC~\citep{Ko2023avdc}, which eliminate action labels through goal-conditioned exploration or dense matching.

As shown in Table~\ref{tab:libero-v2a}, GVF-TAPE achieves the highest performance in 5 tasks and ranks second in the remaining 3. In the 3 tasks, there exists some challenging scenario for our method like robot reach out of camera, we summarize these situation in Fig.~\ref{fig:difficult-scenes}. On average, our overall performance surpasses the second-best approach by 26.9\%. The performance of DP, GCDP, and SuSIE appears relatively low, which may be attributed to their reliance on action-labeled data. Given that only 20 demonstrations per task are available in this experiment, the limited supervision may constrain their effectiveness. These results demonstrate the strong generalization ability of GVF-TAPE across diverse manipulation tasks. Moreover, unlike V2A, which requires costly online exploration for each task, GVF-TAPE operates fully offline by learning from random exploration data, offering improved efficiency and scalability.

Although GVF-TAPE performs robustly across most tasks, some failure cases occur when the end-effector moves outside the camera field of view, particularly in LIVING-ROOM-SCENE-5 (Fig.~\ref{fig:difficult-scenes} (a–b)). Addressing this via multi-view setups or improved pose estimation is left for future work. 

Overall, GVF-TAPE provides a flexible and scalable alternative to video prediction frameworks, fully eliminating the need for action labels or goal-conditioned exploration.

\input{tables/real_world_sr}
\input{images/tex/pretrain_scratch}

\subsection{Real World Performance}
\label{Real World}
We evaluate GVF-TAPE on five real-world tasks involving rigid, deformable, and articulate objects as shown in Fig.~\ref{fig:real_setting}. The tasks include: 1) pick up the blue bowl and place it on the pink plate, 2) grab a tissue, 3) place the sponge on the plate. 4) put the blue bowl into the microwave and close it, and 5) put the pepper in the basket. For each task, we conduct 10 independent trials, with success rates summarized in Tab.~\ref{tab:real_world_sr}. GVF-TAPE achieve an average success rate of 56\% across all tasks with only 20 video per task, using the same task-agnostic pose estimation model without task-specific fine-tuning. Notably, the objects and configurations encountered during evaluation differ from those seen during random exploration training. Despite this domain shift, GVF-TAPE demonstrates strong generalization to unseen object positions, highlighting its robustness and practicality for real-world deployment.
\paragraph{Cross-embodiment Transfer.}

We investigate whether GVF-TAPE can leverage human demonstration videos to improve robot manipulation while reducing reliance on robot-specific data. To this end, we pre-train the video generation module using 50 additional human hand manipulation videos per task, followed by fine-tuning on robot data. This cross-embodiment pre-training improves the model’s ability to capture task-relevant visual structure and generalize across varying spatial configurations. As shown in Fig.~\ref{fig:real_setting}, it reduces hallucinations and enhances real-world robustness. Table~\ref{tab:real_world_sr} reports consistent performance gains, demonstrating that GVF-TAPE can effectively transfer knowledge from human to robot domains and improve generalization with limited robot data.

\input{images/tex/real_setting}

\subsection{Ablation Study}
\label{sec: ablation}

\paragraph{Effect of Rectified Flow.}

To validate our choice of rectified flow~\citep{rectified_flow} for video generation, we compare it with diffusion-based approaches~\citep{ddpm, ddim} used in prior work~\citep{Ko2023avdc, du2023unipi, luo2024groundingv2a}. We evaluate structural similarity (SSIM) and perceptual similarity (LPIPS) across LIBERO-Spatial, LIBERO-Object, and LIBERO-Goal, averaging the metrics across suites (Fig.~\ref{fig:pretrain_scratch_and_ddim_flow}). To accelerate diffusion sampling, we adopt DDIM~\citep{ddim}. As shown, while increasing sampling steps improves diffusion video quality, it significantly increases inference time. In contrast, rectified flow achieves comparable video quality with only three steps, drastically reducing latency. This efficiency is critical for real-time closed-loop deployment. Detailed timing results are provided in the appendix.

\paragraph{Effect of integrating monocular depth estimation.}
We evaluate the impact of incorporating relative depth by comparing GVF-TAPE under two settings: one using RGB-D video generated with supervision from a monocular depth estimator~\citep{video_depth_anything}, and the other using RGB-only video when depth estimation is unavailable. As shown in Table~\ref{tab:rgb_dxr}, integrating depth consistently improves performance across all test environments, with particularly notable gains in spatially complex tasks. Additional experimental results and analyses are included in the supplementary material.
\input{tables/rgb_vs_depth}

%% file: tables/libero_main.tex
\begin{table*}[!ht]
\centering

\setlength{\tabcolsep}{3pt}
\resizebox{1.00\linewidth}{!}{%
\begin{tabular}{l c c c c c c c}
\toprule
Method &Side View &Eye-in-hand View & Action Data & Libero-Spatial & Libero-Object & Libero-Goal & Overall \\

\midrule

R3M-finetune & \Checkmark & \Checkmark & 20\% &\numerr{49.17}{3.79}  &  \numerr{52.83}{8.2}  & \numerr{59.2}{7.80} & \numerr{53.73}{8.04}  \\
VPT & \Checkmark & \Checkmark & 20\%&\numerr{37.83}{4.29}  & \numerr{19.50}{0.82}  & \numerr{3.33}{2.36}  & \numerr{20.22}{14.37}  \\
UniPi & \Checkmark & \Checkmark & 20\% &\underline{\numerr{69.17}{3.75}}  & \numerr{59.83}{3.01}  & \numerr{11.83}{2.02}  & \numerr{46.94}{25.30}  \\
ATM &\Checkmark &  \Checkmark & 20\% &\numerr{68.50}{1.78}  & \underline{\numerr{68.00}{6.18}}  & \textbf{\numerr{77.83}{0.82}}  & \underline{\numerr{71.44}{5.87}}  \\
GVF-TAPE(Ours) & \Checkmark & \XSolidBrush & 0\% &\textbf{\numerr{95.50}{0.87}}   & \textbf{\numerr{86.70}{1.26}}  & \underline{\numerr{66.80}{2.00}}  & \textbf{\numerr{83.00}{12.01}}  \\
\bottomrule
\vspace{-20px}
\end{tabular}

}

\vspace{5pt}
\caption{\footnotesize{ \textbf{Performance comparison with state-of-the-art methods across three LIBERO evaluation suites.}
Success rates (mean ± standard deviation) are reported over three random seeds. GVF-TAPE achieves the highest performance on two of the three suites and outperforms the next-best overall average by 11.56\%. }}

\label{tab:libero-main}
\vspace{-4mm}
\end{table*}

%% file: images/tex/difficult-scenes.tex
\begin{wrapfigure}{r}{0.40\textwidth}
    \centering
    \includegraphics[width=0.98\linewidth, trim=0.0cm 0.0cm 0.0cm 0.3cm,clip]{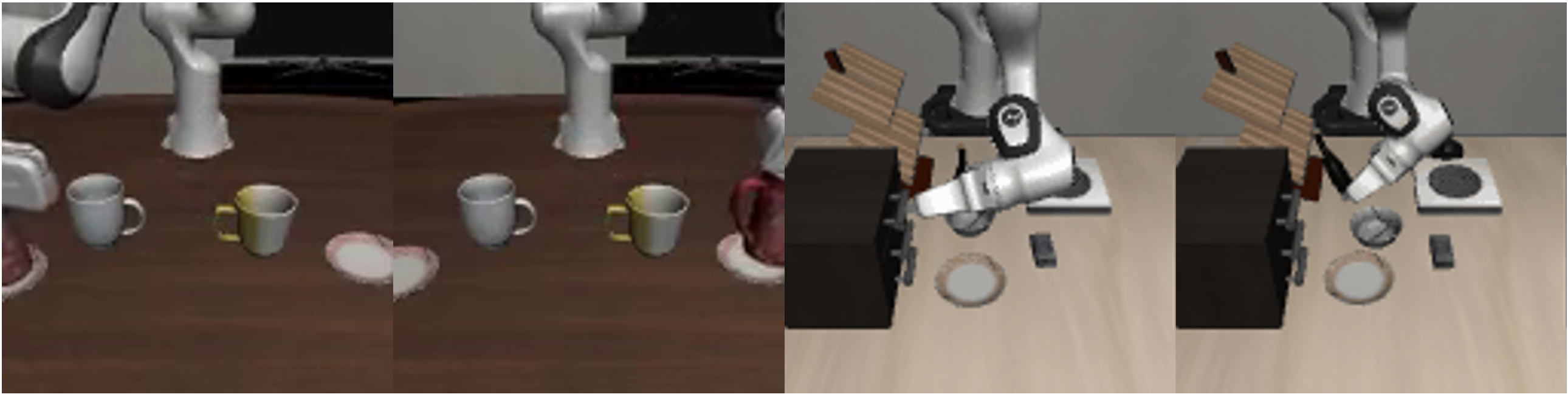}
    \caption{\footnotesize{
    \textbf{Challenging scenarios in LIBERO.}
    The left two panels show tasks from LIVING-ROOM-SCENE-5, where the robot’s end effector moves outside the camera’s field of view, making pose estimation unreliable. The right two panels illustrate limited gripper visibility from a fixed side-view camera, which affects accuracy in fine-grained tasks from LIBERO-Goal.
    }
    }

    \label{fig:difficult-scenes}
    \vspace{-6mm}
\end{wrapfigure}

%% file: images/tex/pretrain_sr.tex
\begin{wrapfigure}{r}{0.38\textwidth}
    \vspace{-5mm} 
    \centering
    \includegraphics[width=0.38\textwidth, trim=0.3cm 0.3cm 0.0cm 1.3cm,clip]{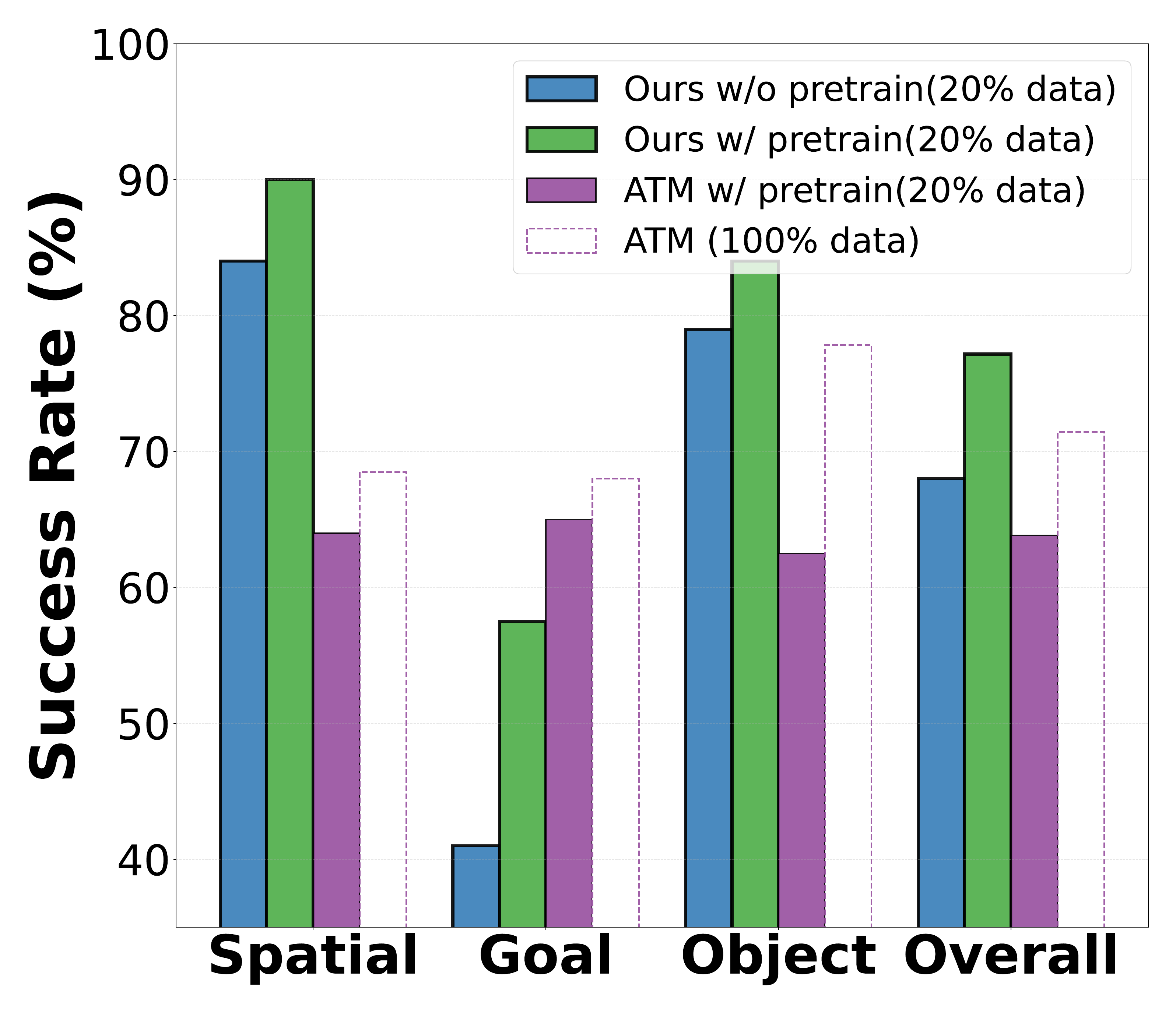}
    \vspace{-5mm}
    \caption{\footnotesize{ \textbf{Performance of our method with and without pretraining.} Using only 20\% of the video data, our method matches prior SOTA (ATM); pretraining on LIBERO-90 boosts performance by 9.2\%, outperforming ATM by 5.43\%.}}
    \label{fig:pre_train_sr}
    \vspace{-3mm}
\end{wrapfigure}

%% file: tables/libero-v2a.tex
\begin{table*}[tbp]
\centering
{
\setlength{\tabcolsep}{3pt}
\resizebox{0.95\linewidth}{!}{%
\begin{tabular}{l c c c c c c c c c}
\toprule
Task & DP* & GCDP* & SuSIE* & AVDC & V2A w/ SuSIE & V2A & Ours \\
\midrule

LR-Scene5-put-red-mug-left & \numerr{33.6}{3.2} & \numerr{24.8}{4.7} & \numerr{18.4}{2.0} & \numerr{0.0}{0.0} & \numerr{23.2}{3.0} & \underline{\numerr{38.4}{15.3}} & \textbf{\numerr{83.6}{6.2}} \\

LR-Scene5-put-red-mug-right & \numerr{33.6}{8.2} & \numerr{22.4}{7.4} & \numerr{32.0}{8.4} & \numerr{0.0}{0.0} & \textbf{\numerr{60.0}{6.7}}   & \numerr{40.8}{7.8}& \underline{\numerr{56.0}{5.7}} \\

LR-Scene5-put-white-mug-left & \numerr{59.2}{7.8} & \numerr{16.0}{8.8} & \numerr{43.2}{4.7} & \numerr{0.0}{0.0} & \textbf{\numerr{68.8}{4.7}} & \numerr{51.2}{3.9} & \underline{\numerr{64.0}{8.0}} \\

LR-Scene5-put-Y/W-mug-right & \numerr{57.6}{5.4} & \numerr{3.2}{3.0} & \numerr{25.6}{11.5} & \numerr{0.0}{0.0} & \textbf{\numerr{67.2}{8.9}} & \numerr{38.4}{8.6} & \underline{\numerr{60.0}{3.6}} \\

LR-Scene6-put-choc-left & \numerr{42.4}{5.4} & \numerr{45.6}{6.0} & \numerr{17.6}{9.3} & \numerr{1.3}{1.9} &  \numerr{44.0}{7.6} & \underline{\numerr{70.4}{12.8}}& \textbf{\numerr{96.8}{1.6}} \\

LR-Scene6-put-choc-right & \numerr{50.4}{5.4} & \numerr{32.0}{8.8} & \numerr{32.8}{9.9} & \numerr{0.0}{0.0} &  \numerr{54.4}{5.4} & \underline{\numerr{79.2}{3.9}}& \textbf{\numerr{92.8}{4.8}} \\

LR-Scene6-put-red-mug-plate & \numerr{32.8}{9.3} & \numerr{5.6}{4.1} & \numerr{16.0}{2.5} & \numerr{0.0}{0.0} & \numerr{66.4}{12.0} & \underline{\numerr{72.8}{6.4}}& \textbf{\numerr{90.4}{3.2}} \\

LR-Scene6-put-white-mug-plate & \underline{\numerr{71.2}{5.3}} & \numerr{7.2}{6.4} & \numerr{10.4}{4.1} & \numerr{0.0}{0.0} & \numerr{36.0}{7.6}& \numerr{25.6}{11.5} & \textbf{\numerr{91.6}{1.7}} \\
\midrule
Overall & \numerr{47.6}{13.4} & \numerr{19.6}{13.7} & \numerr{19.2}{6.5} & \numerr{0.3}{0.5} & \underline{\numerr{52.5}{16.6}}  & \numerr{52.1}{19.7}  & \textbf{\numerr{79.4}{16.6}} \\
\bottomrule
\end{tabular}
}
}
\vspace{-4pt}
\caption{\footnotesize{ 
\textbf{Comparison of methods on eight tasks in two LIBERO-100 living room scenes.} * indicates use of action-label expert demos. Ours outperforms the second-best by 26.9\%.}}

\label{tab:libero-v2a}
\vspace{-3mm}
\end{table*}

%% file: tables/real_world_sr.tex
\begin{wraptable}{r}{0.45\textwidth}
    \vspace{-3mm} 
    \centering
    \setlength{\tabcolsep}{3.0pt}
    \renewcommand{\arraystretch}{0.8}
    \small
    \begin{tabular}{l c c}
        \toprule
        Task & Ours w/o pt. & Ours w/ pt. \\
        \midrule
        put-bowl-plate      & 80\%  & \textbf{100\%} \\
        grab-tissue         & 30\%  & \textbf{70\%}  \\
        put-sponge-plate    & 70\%  & \textbf{90\%}  \\
        bowl-into-micro.    & 60\%  & \textbf{100\%} \\
        pepper-in-basket    & 40\%  & \textbf{70\%}    \\
        \midrule 
        average             & 56\%  & \textbf{86\%}\\
        \bottomrule
    \end{tabular}
    \caption{
    {\small\textbf{Real-world task success rates of our method, with and without pretraining on human hand data.} Pretraining leads to consistently higher performance, reaching 100\% success on several tasks and boosting the overall average by 30\%.}}
    \label{tab:real_world_sr}
    \vspace{-5mm}
\end{wraptable}

%% file: images/tex/pretrain_scratch.tex
\begin{figure}[htbp]
    \centering
    \includegraphics[width=0.9\textwidth,  trim=0.0cm 0.3cm 0.0cm 0.4cm,clip]{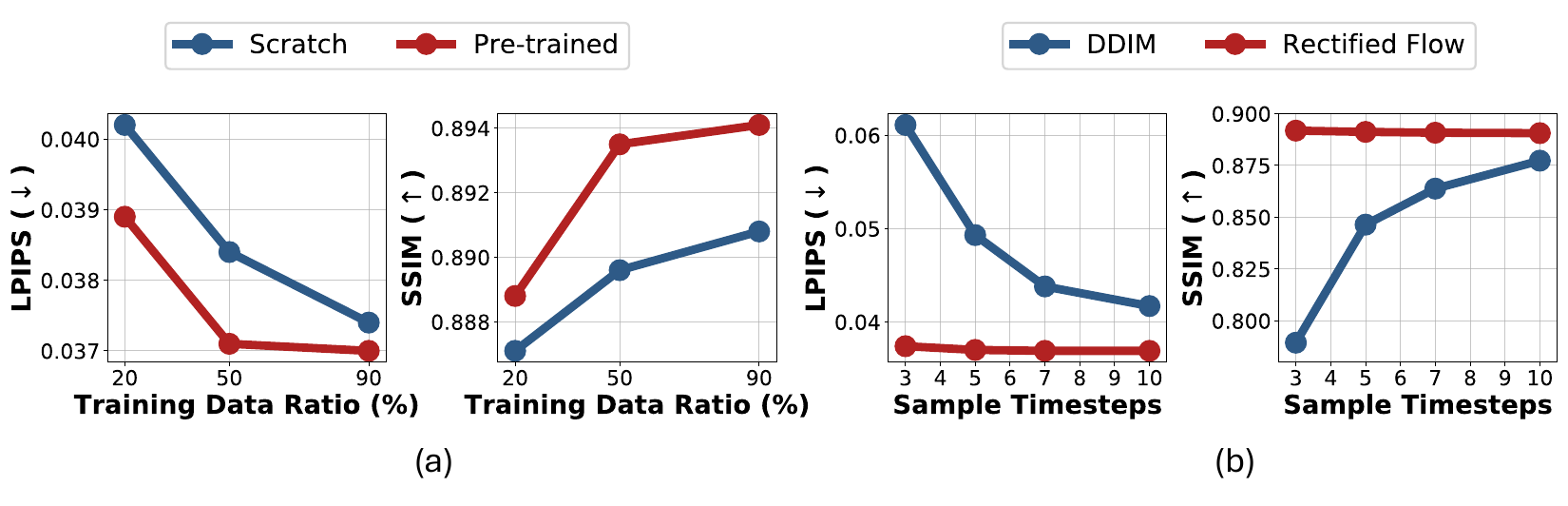}
    \vspace{-2mm}
    \caption{\footnotesize{ \textbf{Pretraining and model choice critically affect video generation quality and efficiency.} (a) Pretrained models consistently outperform models trained from scratch across different proprioception data ratios. (b) While diffusion improves with more sampling steps, it incurs high inference cost; rectified flow achieves strong results with just three steps, motivating our design choice.}}
    
    \label{fig:pretrain_scratch_and_ddim_flow}
    \vspace{-6mm}
\end{figure}

%% file: images/tex/real_setting.tex
\begin{figure}[t]
    \centering
    \includegraphics[width=0.9\columnwidth, trim=0.3cm 0.4cm 0.0cm 0.3cm,clip]{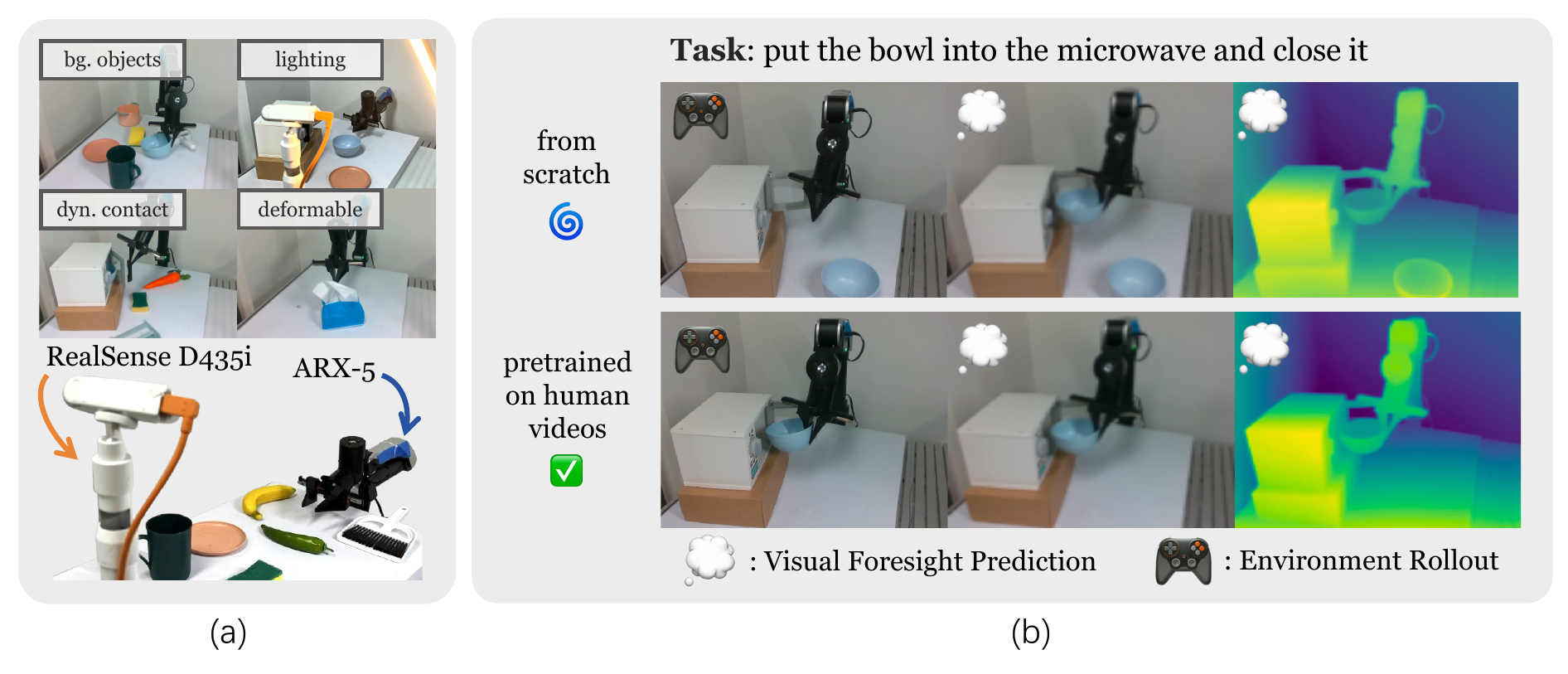}
    \caption{\footnotesize{\textbf{(a) Real-world setup.} We use an ARX-5 robotic arm equipped with a fixed side-view Intel RealSense D435i camera. The evaluation environment includes dynamic contacts, deformable objects, background clutter, and varying lighting conditions.  \textbf{(b) Effect of human video pre-training.} Pre-training on human hand manipulation videos significantly reduces hallucinations and improves prediction stability.}}
    \label{fig:real_setting}
    \vspace{-4mm}
\end{figure}


%% file: tables/rgb_vs_depth.tex
\begin{table*}[!htbp]
\centering
{
\setlength{\tabcolsep}{3pt}
\resizebox{0.85\linewidth}{!}{%
\begin{tabular}{l c c c c c c}
\toprule
Method & Video Depth Anything & Libero-Spatial & Libero-Object & Libero-Goal & Overall \\
\midrule
Ours w/o depth & \XSolidBrush &\numerr{91.83}{1.52}  & \numerr{80.33}{3.33}  & \numerr{56.5}{0.00}  & \numerr{76.22}{14.71}  \\
Ours w/ depth & \Checkmark & \textbf{\numerr{95.50}{0.87}} & \textbf{\numerr{86.70}{1.26}}  & \textbf{\numerr{66.8}{2.00}}  & \textbf{\numerr{83.00}{12.01}} \\
\bottomrule
\end{tabular}
}
}
\vspace{5pt}
\caption{\footnotesize{ \textbf{Performance comparison on three test suites using RGB-D vs. RGB-only input in GVF-TAPE.} Incorporating relative depth significantly boosts performance across all cases, highlighting the benefit of depth information.  }}

\label{tab:rgb_dxr}
\vspace{-7mm}
\end{table*}

%% file: tex/conclusion.tex
\section{Conclusion} 
\label{sec:conclusion}

We present GVF-TAPE, a real-time manipulation framework that decouples visual planning from action execution by combining generative video prediction with task-agnostic pose estimation. Unlike prior methods, GVF-TAPE learns from unlabeled videos and random exploration, removing the need for action-labeled data. This design allows robots to predict future visual outcomes and infer executable poses, enabling robust closed-loop control across diverse tasks. Experiments in both simulation and the real world show that GVF-TAPE outperforms action-supervised and video-based baselines, demonstrating the potential of label-free, foresight-driven frameworks for scalable manipulation. We hope this work encourages further research in video-guided, action-free robot learning.

%% file: tex/future_works.tex
\section{Limitations and Future Works}
While GVF-TAPE achieves strong performance, several limitations remain. First, the system relies exclusively on visual feedback, omitting dynamic signals such as force or tactile feedback that are critical for stable contact-rich manipulation. Incorporating additional sensing modalities, such as proprioception or touch, could improve robustness and interaction awareness. Second, our current single-view video generation model may struggle with fine-grained spatial reasoning due to limited scene coverage. Multi-view foresight could help resolve occlusions and improve accuracy in cluttered or partially observed environments. Finally, although Rectified Flow provides fast and high-quality video prediction, further architectural or optimization improvements could reduce inference latency and enable more agile closed-loop control.

%% file: tex/appendix.tex
\section{Appendix}

\subsection{Performance comparison with VLA methods}
To further evaluate the performance of GVF-TAPE, we compare it with several VLA-based methods, as summarized in Table~\ref{tab:VLA-table}. These baselines are trained with 100\% action-labeled data, while our method uses no action labels. Despite this significant difference in supervision, GVF-TAPE achieves competitive performance, demonstrating the effectiveness of our action label-efficient approach.
\begin{table}[H]
\centering
\renewcommand{\arraystretch}{1.2}
\begin{tabular}{c c c c c}
\toprule
  & Libero-Spatial & Libero-Object & Libero-Goal & Avg.\\
\midrule
Octo\cite{octo_2023} & 78.90 & 85.70 & \textbf{84.60} & 83.07 \\
OpenVLA\cite{kim2024openvla} & 84.70 & 88.40 & 79.20 & 84.10 \\
SpatialVLA\cite{qu2025spatialvla} & 88.20 & \textbf{89.90} & 78.60 & \textbf{85.57} \\
VLA-Cache\cite{xu2025vlacache} & 83.80 & 85.80 & 76.40 & 82.00 \\
TraceVLA\cite{zheng2024tracevla} & 84.60 & 85.20 &75.10 &81.63 \\
GVF-TAPE(ours) & \textbf{95.50} & 86.70 & 66.80 & 83.00\\
\bottomrule
\end{tabular}
\vspace{0.25cm}
\caption{\textbf{Performance comparison with VLA-based methods trained on 100\% action-labeled data.} GVF-TAPE achieves competitive results without requiring action labels, highlighting its label efficiency.}
\label{tab:VLA-table}
\end{table}

\subsection{Overview of the LIBERO benchmark}
\label{sec:overview of libero}

\begin{figure}[H]
    \centering
    \includegraphics[width=0.70\linewidth, trim = 70 160 80 140, clip]{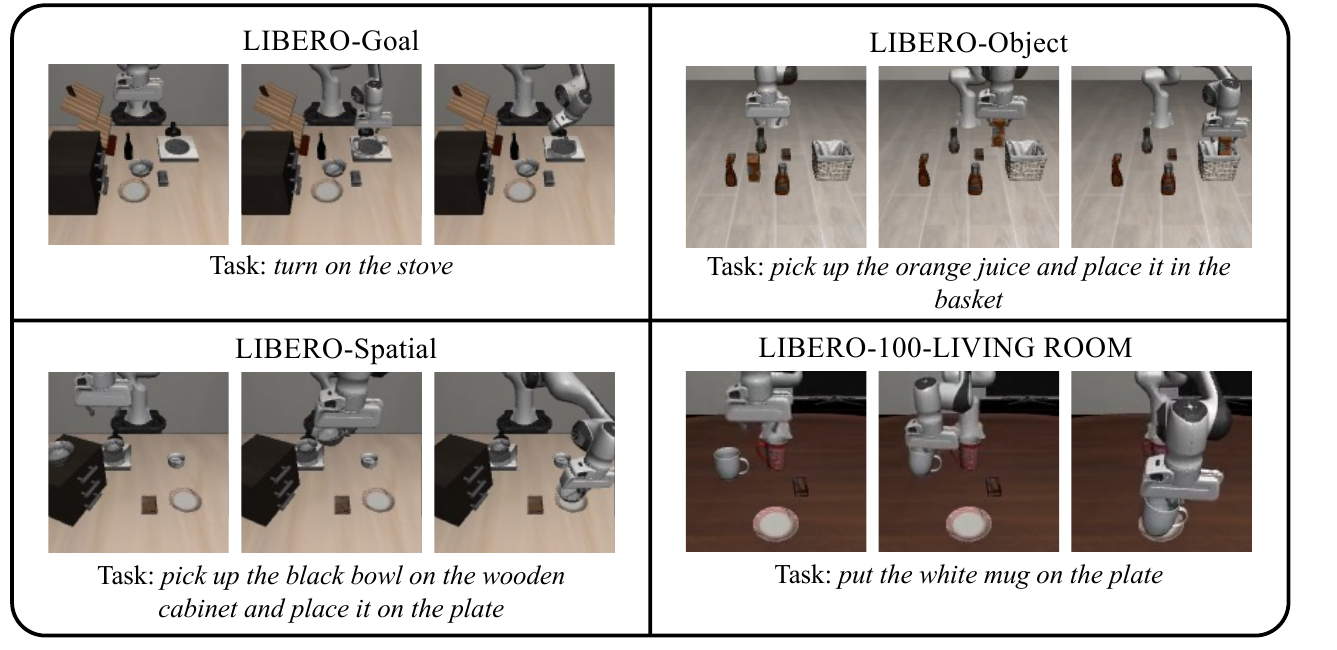}
    \caption{An overview of the LIBERO benchmark.}
    \label{fig:Libero}
\end{figure}

{As illustrated in Fig.~\ref{fig:Libero}, the LIBERO benchmark~\cite{liu2023libero} comprises four task suites: LIBERO-Spatial, LIBERO-Object, LIBERO-Goal, and LIBERO-100. Each of the first three suites contains 10 tasks, while LIBERO-100 includes 100 diverse tasks spanning a wide range of object types and environments. Every task is accompanied by 50 expert demonstrations.

LIBERO-Spatial focuses on spatial variation, such as placing a bowl on a plate at different locations. LIBERO-Object involves manipulating different objects (e.g., pick-and-place tasks), while LIBERO-Goal keeps the object and location fixed but varies the intended goal. LIBERO-100 significantly expands the benchmark with greater diversity in both object types and scene configurations.

The dataset provides side-view and eye-in-hand RGB images at a resolution of 128×128, along with robot proprioception data, supporting both visual and embodied learning tasks.
}

\subsection{Inference Time}

{All real-world evaluations were performed on an NVIDIA RTX 4080 GPU. To improve inference speed, we utilized mixed-precision computation with TensorFloat-32 (TF32) tensor operations. The average computation times are summarized in Table~\ref{tab:inference_time}. GVF-TAPE generates visual plans at an average rate of 1.6 Hz and estimates object poses at 43.5 Hz, as summarized in Table~\ref{tab:inference_time}.}

\input{tables/inference_time}

\input{images/tex/img_encoder}
\subsection{Influences of image encoder structure and modality}
We evaluated several image encoder architectures to assess their impact on pose estimation performance in real-world manipulation tasks. All models were trained on the same dataset, collected via randomized exploration. As shown in Fig.~\ref{fig:img_encoder}, the cross-attention-based encoder consistently outperforms alternative architectures. Specifically, it surpasses a single Vision Transformer (ViT) applied to stacked RGB-D inputs, a standard ViT trained solely on RGB images, and a ResNet-50 backbone. These results suggest that cross-attention mechanisms are particularly effective at integrating and utilizing depth information, making them well-suited for multimodal visual representations in downstream pose estimation tasks.

\subsection{Pose Estimation Model Variants and Implementation}
\label{app: model_var}
To evaluate the impact of different visual encoders on pose estimation performance, we implemented and compared several architectural variants. Each model predicts an 8-dimensional end-effector pose, $T$, based on visual observations, and all models were trained using the same dataset and optimization protocol unless otherwise noted.
\paragraph{Cross-Attention ViT}
Our primary architecture leverages two separate ViT-Base encoders to independently process RGB and depth inputs. These modalities are fused via a multi-head attention mechanism, $\mathcal{A}$, which uses the depth [CLS] token as the query and RGB patch tokens as keys and values. The resulting fused feature is a 768-dimensional vector, passed through a three-layer MLP (256 units per layer) to regress the pose $T$.

\paragraph{Plain RGB ViT}
In this configuration, a single ViT-Base model is used to encode the RGB images. The extracted [CLS] token is input into a three-layer MLP, each with 256 hidden units, to produce the predicted pose. 

\paragraph{Plain RGBD ViT}
Here, the RGB and depth images are concatenated into a four-channel RGB-D input, which is processed by a modified ViT-Base model with a 4-channel input stem. The resulting [CLS] token is used as the feature vector and passed through the same MLP as above. 

\paragraph{ResNet50 Baseline}
 As a convolutional baseline, we use a ResNet-50 model to encode RGB images into a 2048-dimensional vector, which is then mapped to the pose through the same three-layer MLP.

\input{tables/pose_estimation_params}
\input{tables/pose_estimation_simulation}

\subsection{Visual Foresight Model Implementation}
We implement our visual foresight module using a 3D-UNet architecture~\citep{Ko2023avdc} for velocity-based video prediction. To incorporate semantic guidance, we encode textual inputs using CLIP~\citep{clip}, producing latent embeddings that condition the generation process. The 3D-UNet output is modified from 3 to 4 channels to support RGB-D frame generation. This lightweight yet expressive architecture enables the synthesis of spatially-consistent and high-fidelity RGB-D sequences. 

The model is trained using an L2 reconstruction loss, optimized with the AdamW optimizer. We employ a cosine annealing learning rate schedule, starting at $1 \times 10^{-4}$ and decaying to zero over the course of training. Training is performed with a batch size of 8 for 100,000 steps. The model generates 6 future frames per input sequence, and rectified flow fields are computed during inference using an Euler integration scheme. Full implementation details are provided in Table~\ref{tab:visual_foresight}.

\input{tables/visual_foresight}

For pre-training on the Libero-90 dataset, we follow a similar training protocol but increase the batch size to 32 and distribute the training across 4 A100 GPUs in parallel. Pre-training on real-world human hand video data follows the exact procedure outlined in Table~\ref{tab:visual_foresight}.

\subsection{Real World  Experiment Setting}
For real-world experiments, we use an ARX-5 robotic arm paired with a fixed, side-mounted Intel RealSense D435i camera to capture RGB observations. For each manipulation task, we collect 20 teleoperated demonstrations and 50 human hand manipulation videos for pre-training purposes.

The video generation model is trained at a resolution of $128 \times 128$ and conditioned on both the current visual observation and a task-specific language instruction. It generates six future frames per inference. For pose estimation, we collect 18,000 RGB frames (at a resolution of $224 \times 224$) paired with corresponding robot poses using random exploration. Depth maps are labeled using the Video-Depth-Anything model, resulting in an RGB-D and pose dataset.
During real-world evaluation, our system runs using ROS. The generated RGB-D images are resized to $224 \times 224$ and used for pose estimation. The predicted poses are then sent to the ARX-5 robot as control commands through ROS topic. To make the robot’s movements smoother, we employs sinusoidal interpolation for smooth transitions of position and orientation, combined with linear interpolation for the gripper state.

Each task is evaluated over 10 rollouts with varying conditions: 5 with random object placement near the initial position, 2 with objects placed far from the starting location, 2 with added distractor objects on the table, and 1 under altered lighting conditions. Each rollout is limited to 15 video generation cycles; failure to complete the task within this limit is counted as a failure.

\subsection{Simulation Experiment Setting}
For the simulation environments, we use a resolution of 128 for video generation and a resolution of 224 for pose estimation. To ensure the quality of video generation by Video Depth Anything, we rendered the environment at a resolution of 256×256 during random exploration data collection. After generating the depth maps, the data were resized to 224×224 for training the pose estimation network. Our video generation model is trained on the demonstration dataset, while for the pose estimation model, we employ the same random exploration method to collect RGB-D and pose pairs, accumulating over 400,000 such pairs for each suite.

During evaluation, we conduct 20 rollouts for each seed and test a total of 3 evaluation seeds. In the LIBERO environment, we use an additional PID controller to ensure the robotic arm moves to the target pose. Furthermore, to facilitate safe object grasping, we implement a gripper threshold, which set gripper aperture to zero when the generated gripper aperture is below the threshold value.

\begin{figure}[htbp]
    \centering
    \begin{subfigure}[t]{0.48\textwidth}
        \centering
        \includegraphics[width=\textwidth]{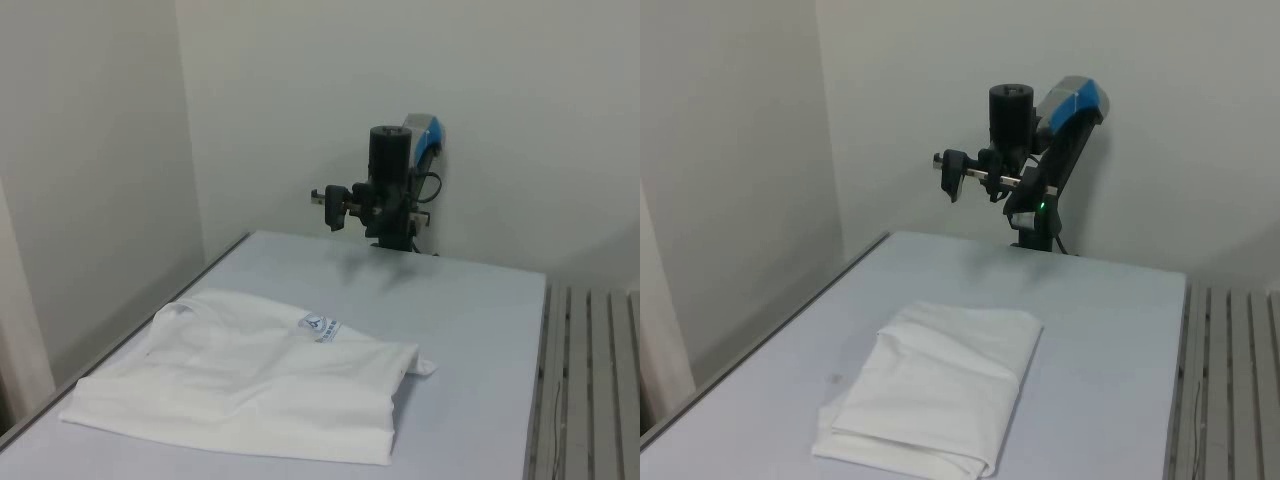}
        \caption{}
        \label{fig:fold_the_cloth}
    \end{subfigure}
    \hfill
    \begin{subfigure}[t]{0.48\textwidth}
        \centering
        \includegraphics[width=\textwidth]{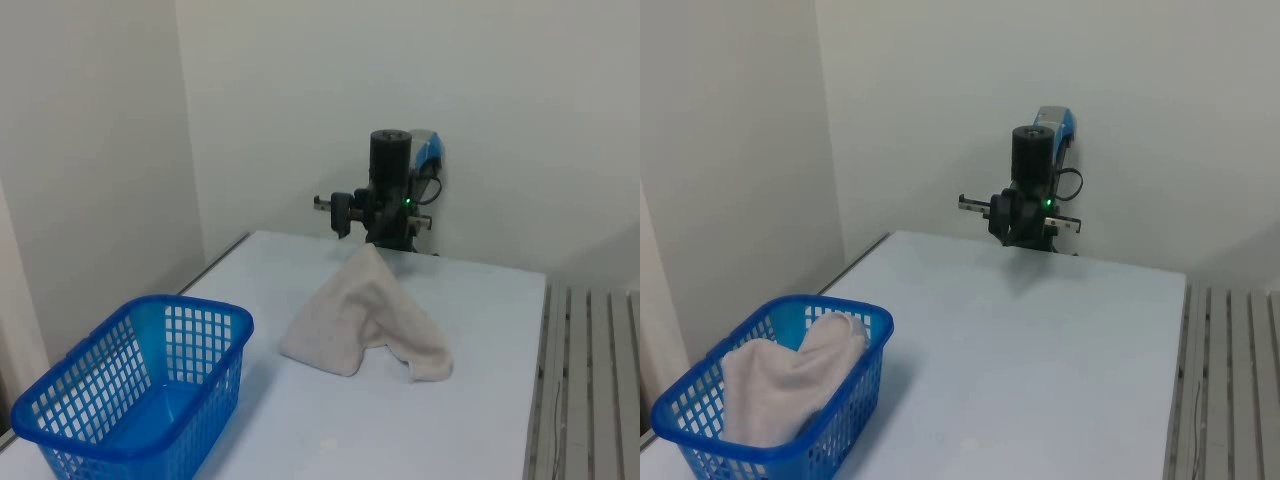}
        \caption{}
        \label{fig:put_the_rag_in_the_trash_bin}
    \end{subfigure}
    \caption{
    \textbf{Real-world task setups for deformable object manipulation.}
\textbf{(a) Fold the cloth:} The robot is required to grasp one edge of the cloth and fold it, and  \textbf{(b) put the rag in the trash bin:} the robot is required to grasp the rag and put it in the trash bin. Each task is shown in its initial and final state. These setups highlight the complexity and variability of real-world deformable object manipulation.}
    \label{fig:deformable_exp.}
\end{figure}

{\subsection{Additional Experiments on Deformable Object Manipulation}
To further evaluate GVF-TAPE’s ability to handle deformable objects, we conducted two additional real-world tasks: (1) folding a cloth, and (2) placing a rag into a trash bin. The task setups and examples of initial and final object states are illustrated in Fig.~\ref{fig:deformable_exp.}. For each task, we collected 20 teleoperated demonstration videos, which were combined with the demonstrations from Section~\ref{Experiment} to train the video generation model. The pose estimation model remained unchanged from Section~\ref{Experiment}. We performed 10 evaluation trials per task, and the results are summarized in Table~\ref{tab:deformable_experiments}.

\begin{table}[H]
\centering
\renewcommand{\arraystretch}{1.2}
\begin{tabular}{l c c}
\toprule
Task & fold the cloth & place-rag-bin \\
\midrule
Success Rate & 70\% & 80\% \\
\bottomrule
\end{tabular}
\vspace{0.25cm}
\caption{\textbf{Success rates for deformable object manipulation tasks.}  GVF-TAPE achieves promising performance on real-world tasks involving deformable objects.}
\label{tab:deformable_experiments}
\end{table}

} 

\subsection{More Qualitative Results}
\label{app: qualitative}

\subsubsection{Failure recovery ability.}
By following video-generation guidance in a closed loop, our system can recover from failures. As shown in Fig.~\ref{fig:failure_recover1}, the video generation model detects the current task state---recognizing that the tissue has not been grasped. After two attempts, it successfully retrieves the tissue using text-prompted instructions.

\input{images/tex/failure_recover}

\subsubsection{Qualitative Comparison of GVF-TAPE w/ and w/o Relative Depth}

The impact of incorporating relative depth is further demonstrated through specific examples comparing the performance of GVF-TAPE under two settings: one using RGB-D video generated with supervision from a monocular depth estimator~\cite{video_depth_anything}, and the other using RGB-only video when depth estimation is unavailable. The inclusion of depth information significantly enhances the system's performance, particularly in spatial pose estimation. Accurate estimation of spatial relationships is critical for successful manipulation. As shown in Fig.\ref{fig:goal_r_tk1_fail}, the RGB-only model produces biased or inaccurate pose estimations, leading to task failure. In contrast, the RGB-D version, demonstrated in Fig.\ref{fig:goal_dxr_tk1_succ_great}, achieves correct pose estimation and successfully completes the tasks.

\input{images/tex/sim_rollouts}
\subsubsection{Failure Analysis}
We summarize several factors contributing to failure, outlined as follows:

\paragraph{Hallucination.}
The video generation model may produce physically implausible frames, such as introducing novel objects or causing the robotic arm to become occluded. As shown in~\ref{fig:hallucination}, the robot may exhibit erratic movement, leading to task failure.

\paragraph{Occultation.}
In certain manipulation tasks, the robotic arm may move behind an object or obstruct its gripper, making pose estimation challenging. As shown in~\ref{fig:occultation}, the proposed method is unable to manipulate the object effectively.

\paragraph{Pose Estimation Error.}
Errors in the pose estimation model can result in incorrect contact positions, preventing the robot from successfully grasping the object. As shown in~\ref{fig:estimation_error}, the robot arm fails to pick up the bowl due to pose estimation error.

To better understand the prevalence of different failure modes, we conducted an evaluation on the LIBERO-Spatial suite. As shown in Tab.~\ref{tab:failure analysis}, out of 200 trials, we observed a total of 11 failures, including 3 due to hallucination, 5 due to pose estimation errors, and 3 due to system-level issues. These results suggest that GVF-TAPE is generally robust in scenarios without significant occlusion. However, its performance may degrade in settings involving occlusion, where accurate pose estimation becomes more difficult.

\begin{table}[H]
\centering
\renewcommand{\arraystretch}{1.2}
\begin{tabular}{c c c c c}
\toprule
Total Trials & Success & Pose Est. Error & Hallucination & Sys.-Level Error \\
\midrule
200 & 189 & 5 & 3 & 3 \\
\bottomrule
\end{tabular}
\vspace{0.25cm}
\caption{\textbf{Failure Analysis}}
\label{tab:failure analysis}
\end{table}

\input{images/tex/failure_analysis}

\subsection{Random Exploration}
\label{app:random_exploration}
{
The random exploration process employs a randomized sampling strategy to acquire diverse end-effector poses within the robot's operational workspace and within FOV of the agentview camera. 

In real-world settings, to ensure safety, we incorporate several safeguards, including joint limit checks and unexpected stop detection. The entire sampling process runs autonomously at 10~Hz, enabling stable and continuous operation.

This approach enables efficient exploration of the reachable workspace while maintaining continuous operation stability. In real world settings, we collect around 18k pose-image pair data. The pseudo code real world sampling strategy of our method is
provided in Algorithm \ref{alg:AOS}.
}

\begin{algorithm}[!h]
\caption{Random Exploration Algorithm in the Real World}
\label{alg:AOS}
\begin{algorithmic}[1]
\REQUIRE Workspace bounds $\mathcal{W}$, arrival threshold $\Delta\mathcal{T}$, number of frames $N$
    \STATE \textbf{Start a parallel thread to continuously check safety}
    \WHILE{$num_{frames} < N$}
        \STATE Sample a desired end-effector pose $p_{desire} \in \mathcal{W}$
        \WHILE{$\|p_{current} - p_{desire}\|_2 < \Delta\mathcal{T}$} 
            \STATE Resample $p_{desire}$
            \STATE Set $p_{desire}$ as the new goal and publish to the robot arm controller
            \IF{$p_{current} \notin \mathcal{W}$}
                \STATE Resample $p_{desire} \in \mathcal{W}$ and publish to the robot arm controller
            \ENDIF
        \ENDWHILE
    \ENDWHILE
\end{algorithmic}
\end{algorithm}

\subsubsection{Qualitative Results of Real World Tasks}

The following are visualizations of real-world tasks: 1) pick up the blue bowl and place it on the pink plate ~\ref{fig:blue_bowl_succ}, 2) grab a tissue~\ref{fig:tissue_succ}, 3) place the sponge on the plate~\ref{fig:sponge_succ}. 4) put the blue bowl into the microwave and close it~\ref{fig:microwave_succ}, and 5) put the pepper in the basket~\ref{fig:pepper_success}. 6)fold the cloth~\ref{fig:fold the cloth}. 7) put the rag in the trash bin~\ref{fig:put the rag in the trash}.

\input{images/tex/real_world_rollouts}

%% file: tables/inference_time.tex
\begin{table}[H]
\centering
\renewcommand{\arraystretch}{1.2}
\caption{\textbf{Inference time.}}
\label{tab:inference_time}
\begin{tabular}{l c c  }
\toprule
\textbf{Module} & Video Generation & Pose Estimation \\
\midrule
\textbf{Cost Time (s)} &\numerr{0.61}{0.0037} & \numerr{0.023}{0.0096} \\

\bottomrule
\end{tabular}
\end{table}  

%% file: images/tex/img_encoder.tex
\begin{figure}[htb]
    \vspace{2mm}
    \centering
    \includegraphics[width=0.98\linewidth]{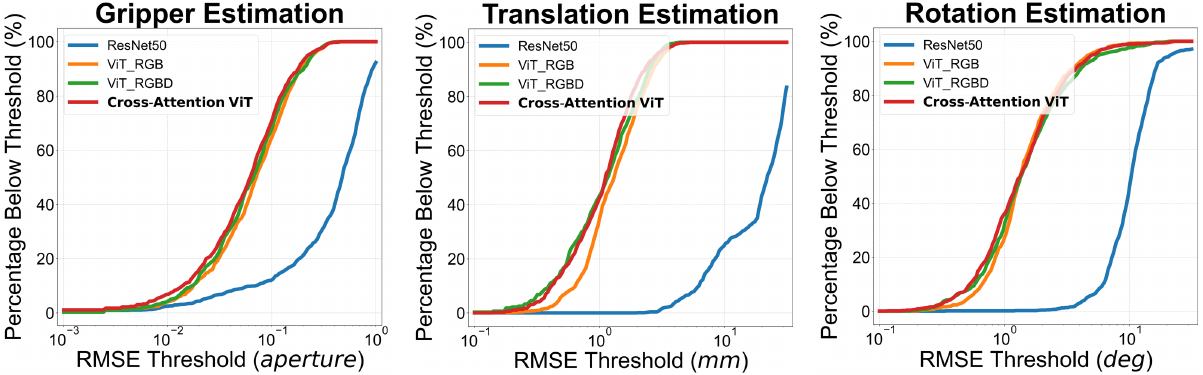}
    \label{fig:comparison}
    \caption{\small  \textbf{Comparison of model architecture.} Performance evaluation using AUC for ResNet50, RGB 3-channel ViT, RGBD 4-channel ViT, and Depth-RGB cross attention model as pose estimation network, trained separately on same amount of random exploration data. Each point on the curve represents the percentage of test points within a given threshold, with a larger AUC indicating better performance. }

    \label{fig:img_encoder}
    \vspace{-4mm}
\end{figure}

%% file: tables/pose_estimation_params.tex
\begin{table}[H]
\centering
\renewcommand{\arraystretch}{1.2}
\caption{Pose Estimation Model and Training Parameters for real world experiment}
\label{tab:pose_estimation_params}
\begin{tabular}{l c}
\toprule
\textbf{Parameter} & \textbf{Value} \\
\midrule
Activation Function & ReLU \\

Optimizer & Adam \\
Learning Rate & \(1 \times 10^{-4}\) \\
\(\beta\) Values & [0.9, 0.999] \\
Weight Decay & \(1 \times 10^{-8}\) \\
Epochs & 100 \\
Batch Size & 512 \\
Color Jitter (B, C, S, H) & (0.3, 0.2, 0.3, 0.2) \\
Image Resize (H, W) & (224, 224) \\
Normalize RGB (Mean, Std) & ([0.5, 0.5, 0.5], [0.5, 0.5, 0.5]) \\
Normalize Depth (Mean, Std) & ([0.5], [0.5]) \\
\bottomrule
\end{tabular}
\end{table}

%% file: tables/pose_estimation_simulation.tex
\begin{table}[H]
\centering
\renewcommand{\arraystretch}{1.2}
\caption{Pose Estimation Model and Training Parameters for Simulation}
\label{tab:pose_estimation_simulation}
\begin{tabular}{l c}
\toprule
\textbf{Parameter} & \textbf{Value} \\
\midrule
Activation Function & ReLU \\

Optimizer & Adam \\
Learning Rate & \(1 \times 10^{-4}\) \\
\(\beta\) Values & [0.9, 0.999] \\
Weight Decay & \(1 \times 10^{-8}\) \\
Epochs & 100 \\
Batch Size & 128 \\
Color Jitter (B, C, S, H) & (0.3, 0.2, 0.3, 0.2) \\
Image Resize (H, W) & (224, 224) \\
Normalize RGB (Mean, Std) & ([0.5, 0.5, 0.5], [0.5, 0.5, 0.5]) \\
Normalize Depth (Mean, Std) & ([0.5], [0.5]) \\
\bottomrule
\end{tabular}
\end{table}

%% file: tables/visual_foresight.tex
\begin{table}[H]
\centering
\renewcommand{\arraystretch}{1.2}
\caption{Visual Foresight Model Implementation}
\label{tab:visual_foresight}
\begin{tabular}{l c}
\toprule
\textbf{Parameter} & \textbf{Value} \\
\midrule
Loss Function & L2 \\

Optimizer & AdamW \\
LR Scheduler & CosineAnnealing \\
Init Learning Late & \(1 \times 10^{-4}\) \\
Weight Decay & \(1 \times 10^{-8}\) \\
Decay Period & 100000\\
Steps & 100000 \\
Batch Size & 8 \\
Generation Frames & 6 \\
Rectified Flow Solver & Euler Solver \\

\bottomrule
\end{tabular}
\end{table}

%% file: images/tex/failure_recover.tex
\begin{figure*}[htbp]
    \centering
    \includegraphics[width=0.85\linewidth]{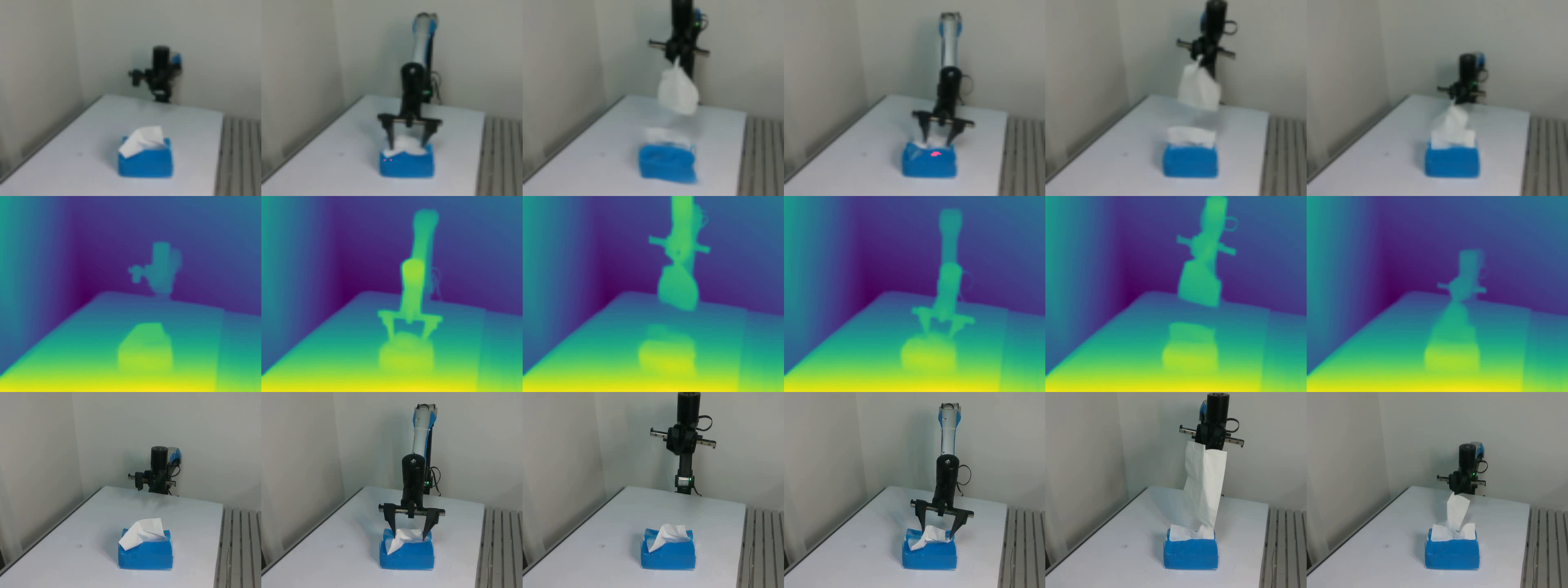}
    \caption{Eval environment roll out of successfully grabbing a tissue through multiple replans. The first and second rows show generated RGB and depth frames, respectively; the third row shows the real world environment. The robot arm fail to grab out the tissue during the first trial; Video generation model as a planner in this process notice the tissue hasn't been grabbed, so the new sampled image will still direct the robot to do so, leading the final success.}
    \label{fig:failure_recover1}
\end{figure*}


%% file: images/tex/sim_rollouts.tex

\begin{figure*}[htbp]
    \centering
    \includegraphics[width=0.85\linewidth]{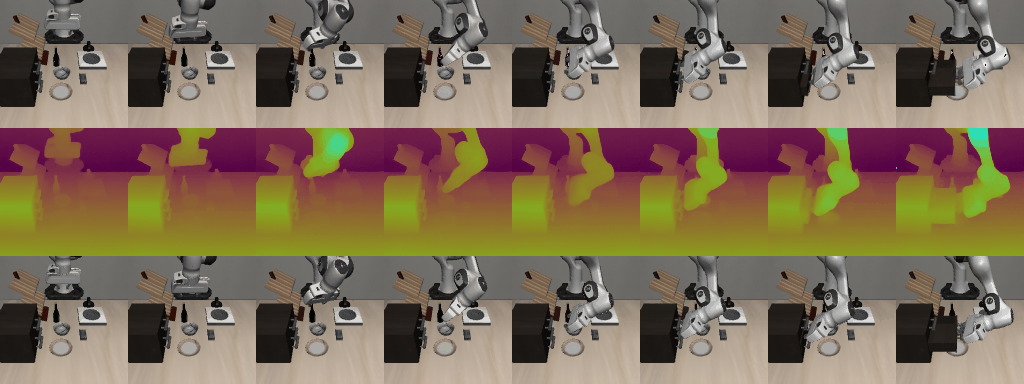}
    \caption{Evaluation rollout of the system with Video-Depth-Anything successfully opening the drawer. The first and second rows show generated RGB and depth frames, respectively; the third row shows the simulation environment.}
    \label{fig:goal_dxr_tk1_succ_great}
\end{figure*}

\begin{figure*}[htbp]
    \centering
    \includegraphics[width=0.85\linewidth]{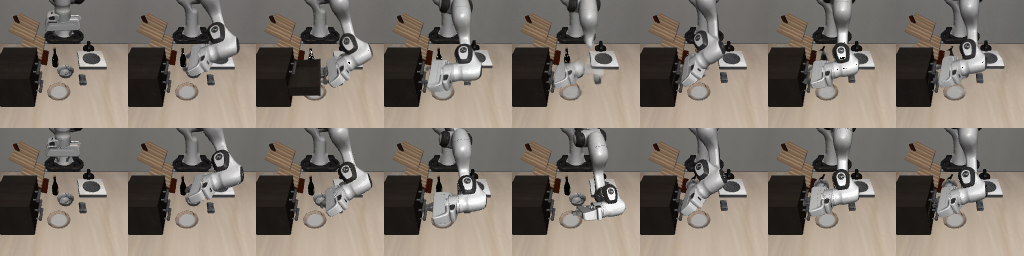}
    \caption{Evaluation rollout of the system without Video-Depth-Anything failing to open the drawer due to biased spatial pose estimation. The first row shows generated RGB frames; the second row shows the simulation environment.}
    \label{fig:goal_r_tk1_fail}
\end{figure*}

%% file: images/tex/failure_analysis.tex

\begin{figure*}[htbp]
    \centering
    \includegraphics[width=0.85\linewidth]{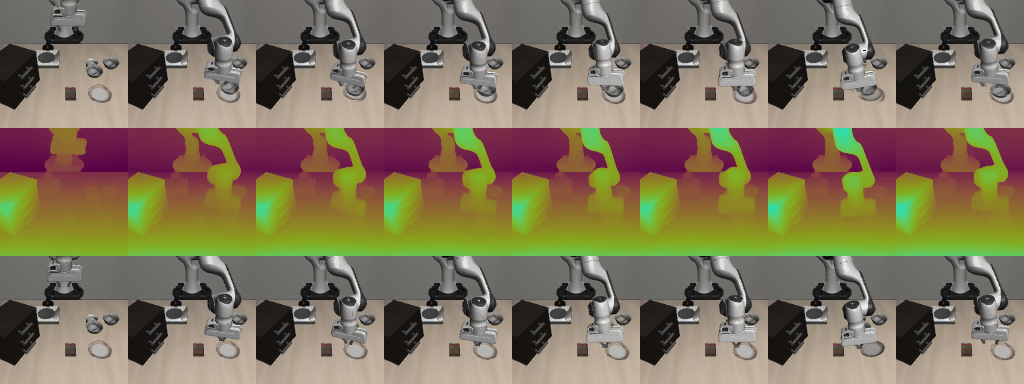}
    \caption{Hallucination in the video generation model leads to task failure. The figure above illustrates a scenario where the model generates a novel bowl, resulting in failure to complete the task. The first and second rows display the generated RGB and depth frames, respectively, while the third row depicts the simulation environment.}
    \label{fig:hallucination}
\end{figure*}

\begin{figure*}[htbp]
    \centering
    \includegraphics[width=0.85\linewidth]{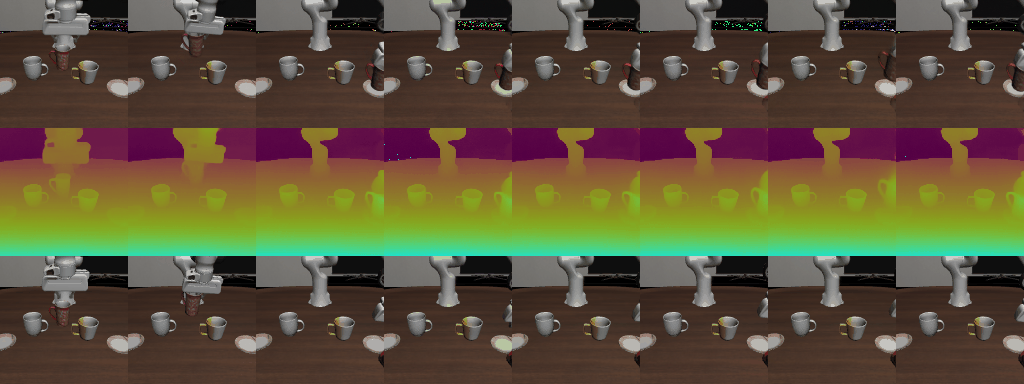}
    \caption{Occultation of the gripper leads to failure. The figure above demonstrates a scenario where the robotic arm moves out of the camera's view, resulting in unreliable pose estimation.  The first and second rows display the generated RGB and depth frames, respectively, while the third row depicts the simulation environment.}
    \label{fig:occultation}
\end{figure*}
\begin{figure*}[htbp]
    \centering
    \includegraphics[width=0.85\linewidth]{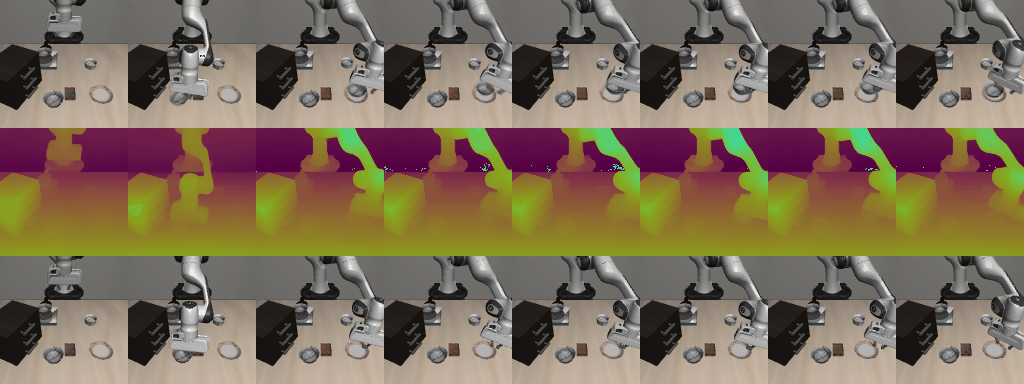}
    \caption{Pose estimation errors lead to failure. The figure above illustrates a scenario where the robotic arm fails to grasp the bowl due to inaccurate pose estimation.  The first and second rows display the generated RGB and depth frames, respectively, while the third row depicts the simulation environment.}
    \label{fig:estimation_error}
\end{figure*}

%% file: images/tex/real_world_rollouts.tex
\begin{figure*}[htbp]
    \centering
    \includegraphics[width=0.9\linewidth]{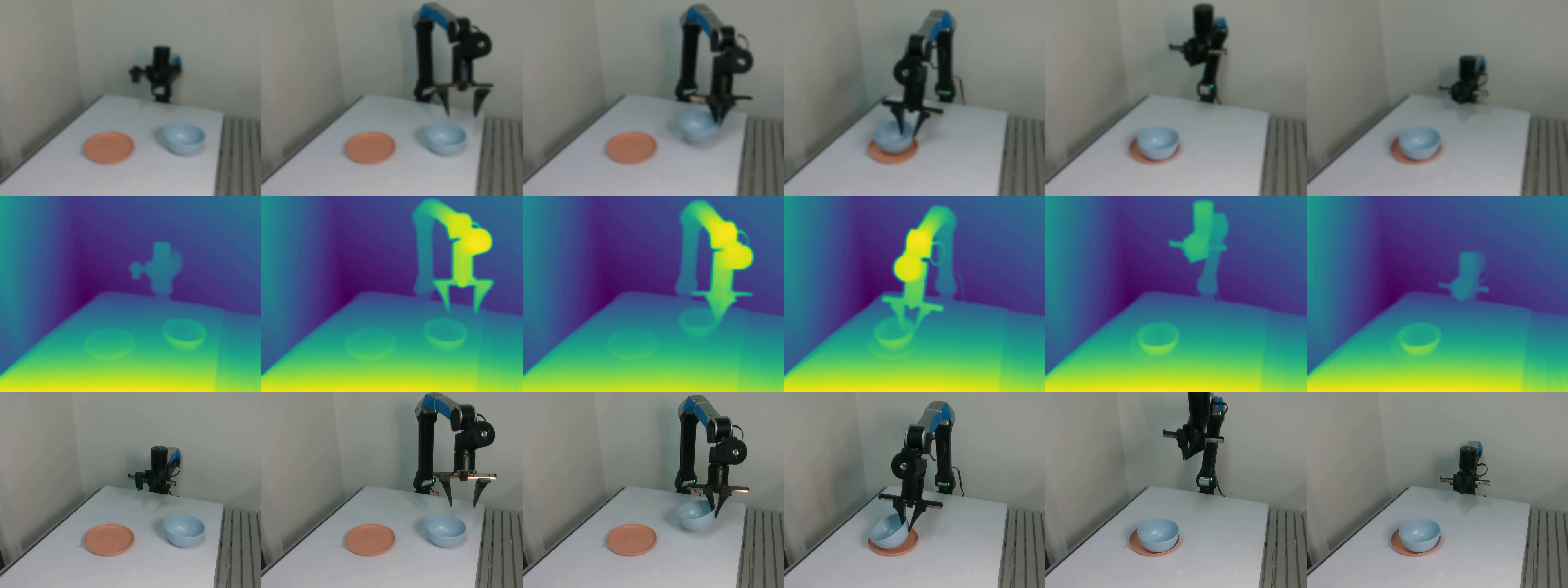}
    \caption{\textbf{Evaluation rollout of real world task pick up the blue bowl and place it on the pink plate.} The first and second rows show generated RGB and depth frames, respectively; the third row shows the real world environment.}
    \label{fig:blue_bowl_succ}
\end{figure*}

\begin{figure*}[htbp]
    \centering
    \includegraphics[width=0.9\linewidth]{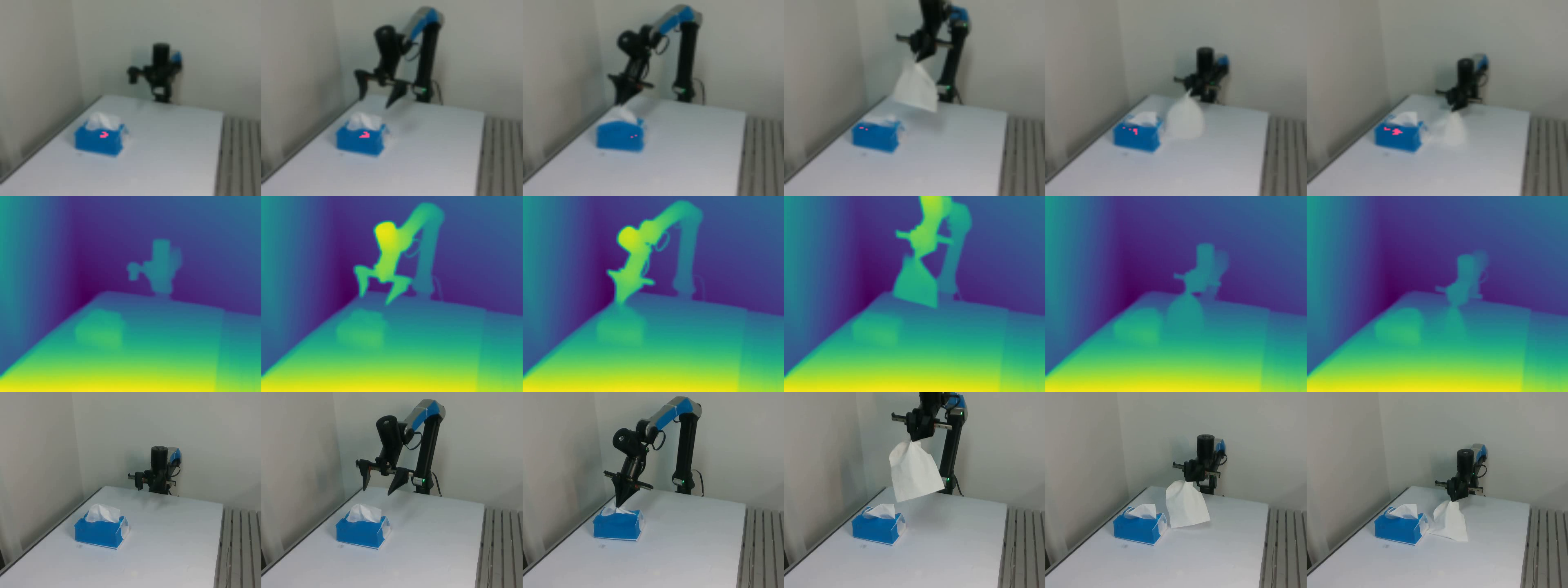}
    \caption{\textbf{Evaluation rollout of real world task grab a tissue.} The first and second rows show generated RGB and depth frames, respectively; the third row shows the real world environment.}
    \label{fig:tissue_succ}
\end{figure*}

\begin{figure*}[htbp]
    \centering
    \includegraphics[width=0.9\linewidth]{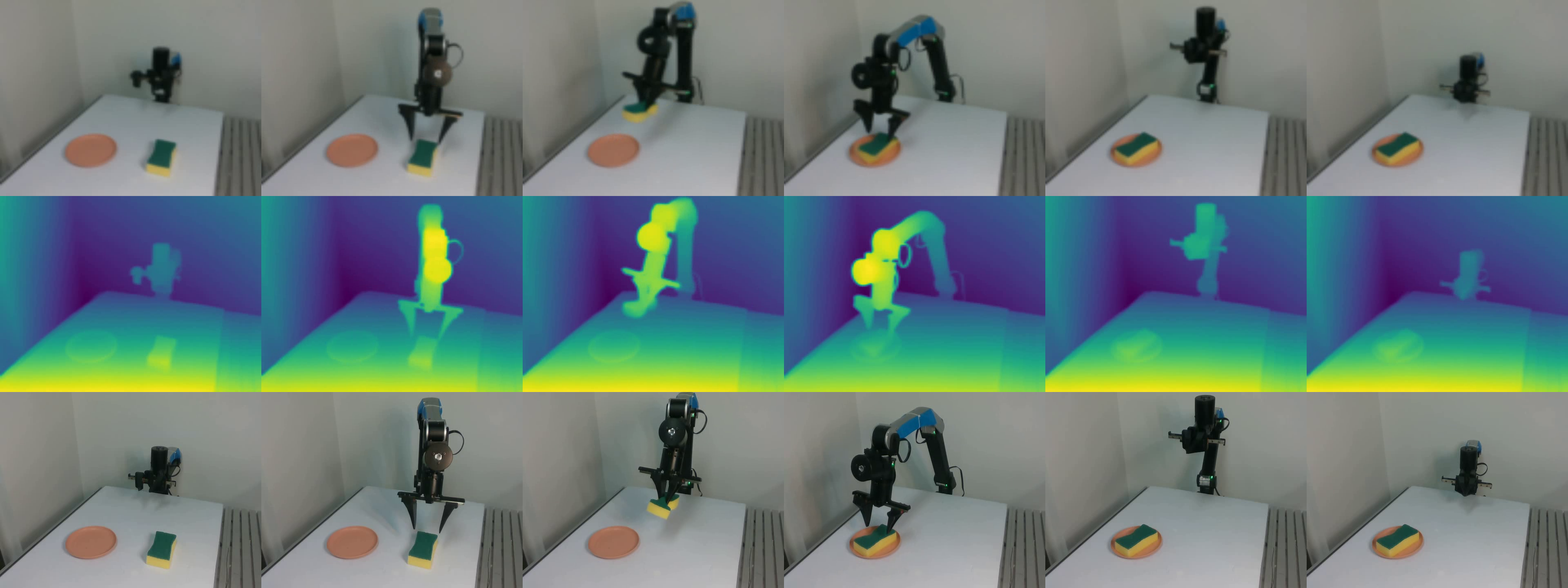}
    \caption{\textbf{Evaluation rollout of real world task place the sponge on the plate.} The first and second rows show generated RGB and depth frames, respectively; the third row shows the real world environment.}
    \label{fig:sponge_succ}
\end{figure*}

\begin{figure*}[htbp]
    \centering
    \includegraphics[width=0.9\linewidth]{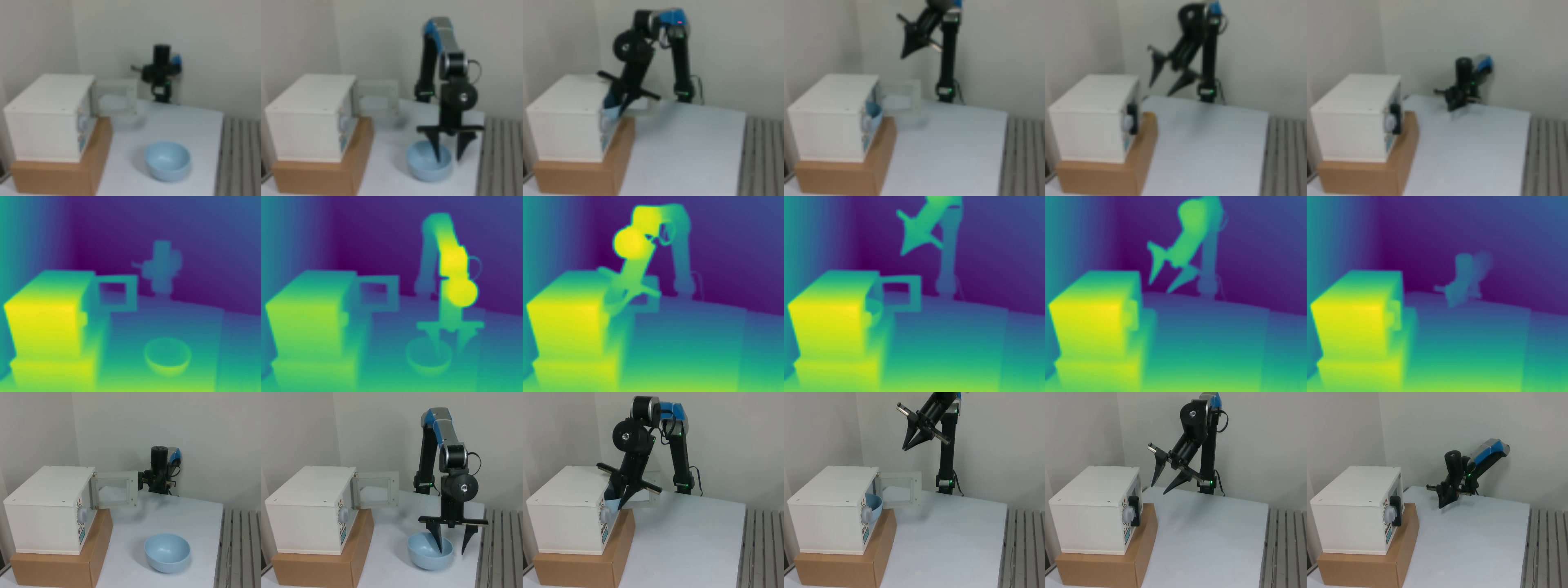}
    \caption{\textbf{Evaluation rollout of real world task put the blue bowl into the microwave and close it.} The first and second rows show generated RGB and depth frames, respectively; the third row shows the real world environment.}
    \label{fig:microwave_succ}
\end{figure*}

\begin{figure*}[htbp]
    \centering
    \includegraphics[width=0.9\linewidth]{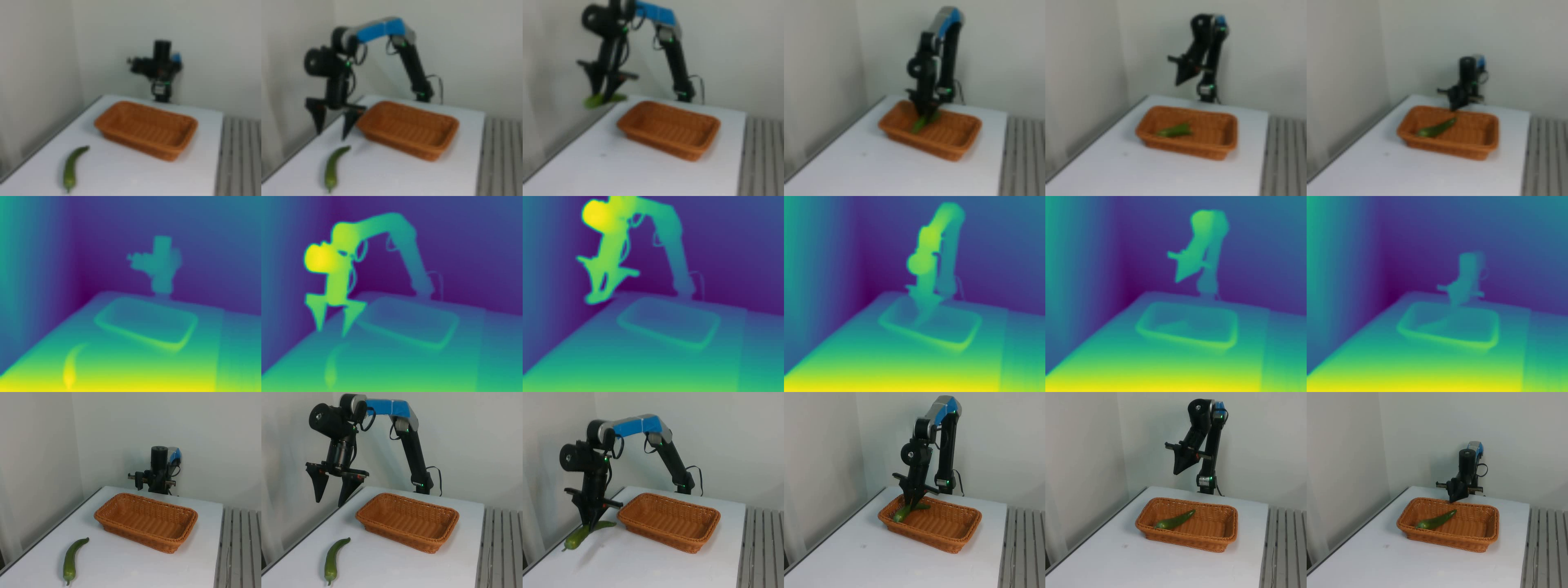}
    \caption{\textbf{Evaluation rollout of real world task put the pepper in the basket.} The first and second rows show generated RGB and depth frames, respectively; the third row shows the real world environment.}
    \label{fig:pepper_success}
\end{figure*}

\begin{figure*}[htbp]
    \centering
    \includegraphics[width=0.9\linewidth]{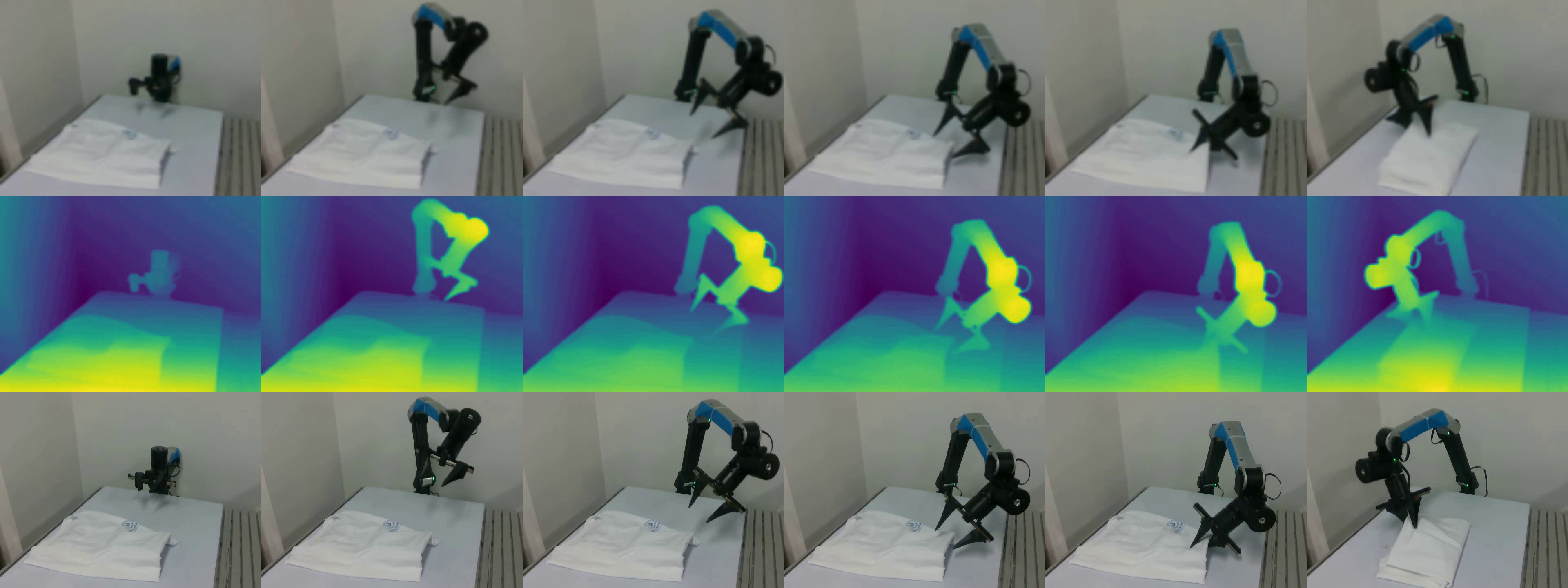}
    \caption{\textbf{Evaluation rollout of real world task fold the cloth.} The first and second rows show generated RGB and depth frames, respectively; the third row shows the real world environment.}
    \label{fig:fold the cloth}
\end{figure*}

\begin{figure*}[htbp]
    \centering
    \includegraphics[width=0.9\linewidth]{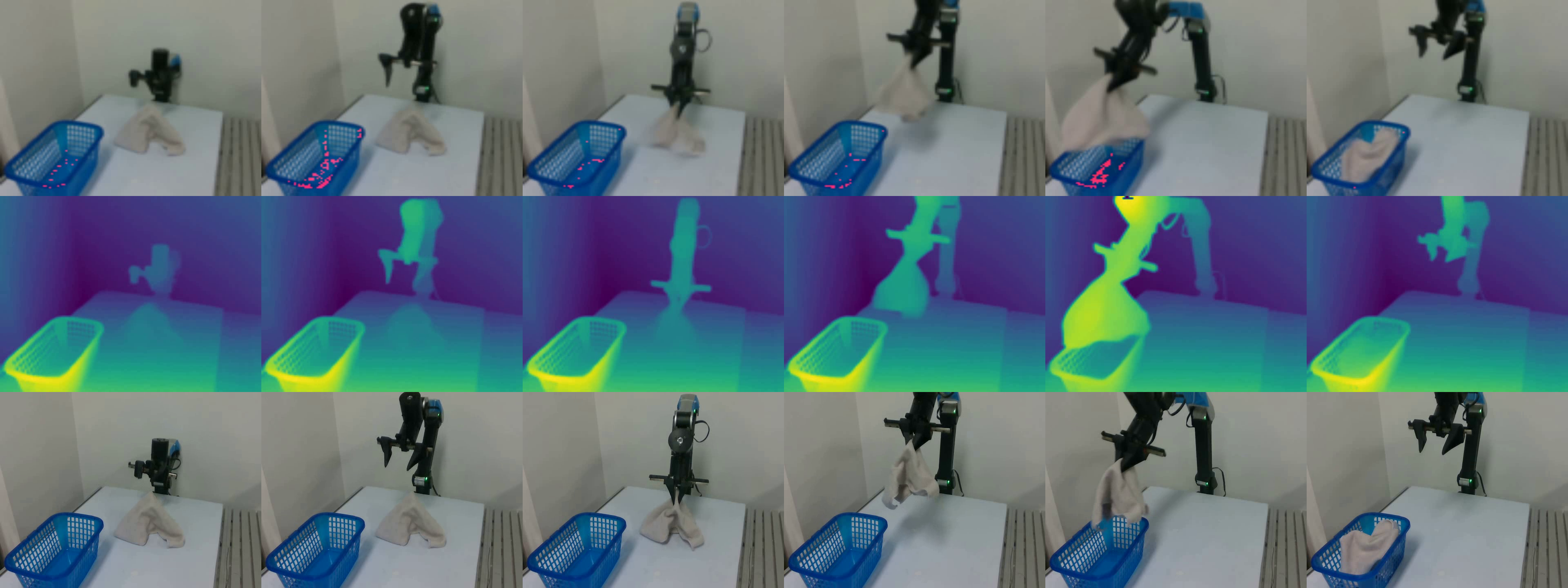}
    \caption{\textbf{Evaluation rollout of real world task put the rag in the trash bin.} The first and second rows show generated RGB and depth frames, respectively; the third row shows the real world environment.}
    \label{fig:put the rag in the trash}
\end{figure*}